\documentclass[final,5p,times,twocolumn]{elsarticle}
\usepackage[disable]{endfloat}

\usepackage{amssymb}
\usepackage{amsmath}
\usepackage{amsfonts}
\usepackage{algorithm}
\usepackage{algorithmic}
\usepackage{multirow}
\usepackage{url}
\usepackage{xcolor}
\usepackage{verbatim}

\journal{}

\begin{document}

\begin{frontmatter}



\title{Modeling Higher-Order Brain Interactions via a Multi-View Information Bottleneck Framework for fMRI-based Psychiatric Diagnosis}


\author[label1]{Kunyu Zhang}
\author[label2]{Qiang Li\corref{cor1}}\ead{qli27@gsu.edu}
\author[label2]{Vince D. Calhoun}
\author[label1,label3]{Shujian Yu\corref{cor1}}\ead{s.yu3@vu.nl}
\cortext[cor1]{Corresponding authors.}

\affiliation[label1]{organization={Department of Computer Science, Vrije Universiteit Amsterdam},
            city={Amsterdam},
            postcode={1081 HV},
            country={The Netherlands}}

\affiliation[label2]{organization={Tri-Institutional Center for Translational Research in Neuroimaging and Data Science (TReNDS), Georgia State University, Georgia Institute of Technology, and Emory University},
            city={Atlanta},
            state={GA},
            postcode={30303},
            country={USA}}

\affiliation[label3]{organization={Department of Physics and Technology, UiT - The Arctic University of Norway},
            city={Tromsø},
            postcode={9019},
            country={Norway}}

\begin{abstract}
Resting-state functional magnetic resonance imaging (fMRI) has emerged as a cornerstone for psychiatric diagnosis, yet most approaches rely on pairwise brain cortical or sub-cortical connectivities that overlooks higher-order interactions (HOIs) central to complex brain dynamics. While hypergraph methods encode HOIs through predefined hyperedges, their construction typically relies on heuristic similarity metrics and does not explicitly characterize whether interactions are synergy- or redundancy-dominated. 
In this paper, we introduce $O$-information, a signed measure that characterizes the informational nature of HOIs, and integrate third- and fourth-order $O$-information into a unified multi-view information bottleneck framework for fMRI-based psychiatric diagnosis. To enable scalable $O$-information estimation, we further develop two independent acceleration strategies: a Gaussian analytical approximation and a randomized matrix-based R\'enyi entropy estimator, achieving over a 30-fold computational speedup compared with conventional estimators.
Our tri-view architecture systematically fuses pairwise, triadic, and tetradic brain interactions, capturing comprehensive brain connectivity while explicitly penalizing redundancy. Extensive evaluation across four benchmark datasets (REST-meta-MDD, ABIDE, UCLA, ADNI) demonstrates consistent improvements, outperforming 11 baseline methods including state-of-the-art graph neural network (GNN) and hypergraph based approaches. Moreover, our method reveals interpretable region-level synergy-redundancy patterns which are not explicitly characterized by conventional hypergraph formulations.
\end{abstract}



\begin{keyword}
Multi-view learning \sep Information Bottleneck \sep $O$-information \sep Psychiatric diagnosis \sep Resting-state fMRI \sep Brain network analysis



\end{keyword}

\end{frontmatter}



\section{Introduction}
Mental disorders exhibit complex and distributed neural signatures, making precise neurobiological characterization challenging. Resting-state functional magnetic resonance imaging (rs-fMRI)~\cite{matthews2004fmri} has become a cornerstone for machine learning–based diagnostic frameworks, as it enables noninvasive assessment of whole-brain functional network connectivity (FNC) alterations across diverse patient populations~\cite{biswal2010discovery, calhoun2007schizophrenia, yang2024brainmass}. Earlier machine learning pipelines rely on manually crafted graph-theoretic features~\cite{shen2017using}, which can be ineffective or biased. Deep learning approaches, including CNNs~\cite{lin2022sspnet} and adversarial learning~\cite{tang2023multisite}, mitigate such biases by learning hierarchical representations directly from data, yet they may not fully leverage the non-Euclidean structure of brain networks. Graph neural networks (GNNs)~\cite{kipf2016semi} address this gap by modeling brain networks as graphs with regions of interest (ROIs) or independent component networks (ICNs) as nodes and FNC as edges~\cite{iraji2023identifying, lei2022graph}, achieving promising diagnostic accuracy on autism spectrum disorder (ASD)~\cite{li2021braingnn, tang2023multisite}, major depressive disorder (MDD), schizophrenia (SZ)~\cite{gaussOinfo2310, lei2022graph}, and Alzheimer's disease (AD)~\cite{zhang2026pime}.

Yet two fundamental limitations of GNN-based pipelines persist. First, most pipelines quantify FNC through correlations or partial correlations, thereby restricting inference to linear, pairwise dependencies and overlooking higher-order interactions (HOIs) that are central to complex cognition~\cite{prado2022neurobiology}. Second, they are sensitive to noisy or spurious interactions arising from measurement noise or subject-specific artifacts~\cite{power2012spurious}, which can undermine generalization and model reliability~\cite{wang2026usbd}.

\begin{figure}[t]
  \centering
  \includegraphics[width=0.9\linewidth]{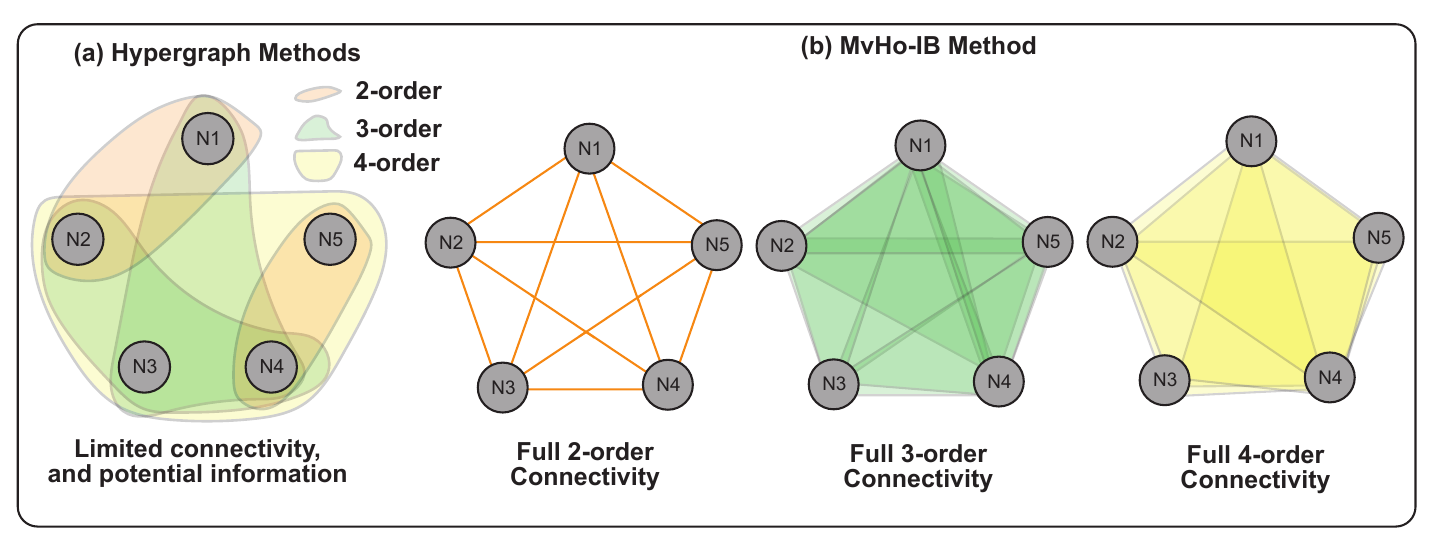}
  \caption{Comparison between existing hypergraph methods and our MvHo-IB++: (a) hypergraph methods represent HOIs through selected hyperedges, often defined by predefined criteria. (b) MvHo-IB++ instead exhaustively models pairwise, third-order, and fourth-order interactions, while explicitly distinguishing synergy-dominated from redundancy-dominated patterns through $O$-information. The IB regularization further compresses view-specific redundant information in the learned representations.}
  \label{fig:compare_hypergraph}
\end{figure}

Hypergraph formulations address the first limitation by enabling edges that bind arbitrary-size sets of ROIs, thus explicitly encoding multivariate functional dependencies~\cite{feng2019hypergraphneuralnetworks, HL}. However, the construction of hypergraphs typically relies on manually chosen similarity metrics and pruning rules~\cite{cai2022hypergraph}, which may introduce selection bias. The resulting hyperedges indicate co-association without revealing \emph{how} regions share information, e.g., whether an interaction is redundancy- or synergy-dominated~\cite{scagliarini2024gradients}. From an information-theoretic perspective, $\mathcal{O}$-information~\cite{rosas2019oinfo} offers a data-driven information-theoretic alternative: it provides a signed multivariate dependence measure where negative values indicate synergy-dominated interactions generating genuinely new joint information and positive values indicate redundancy-dominated interactions reflecting repeated signals. As illustrated in Figure~\ref{fig:compare_hypergraph}, $O$-information captures multiregion dependencies and characterizes whether they are synergy- or redundancy-dominated, thereby supporting the modeling of both pairwise and higher-order interactions.

Despite its conceptual appeal, a rigorous implementation remains challenging, particularly with respect to the efficient estimation of $\mathcal{O}$-information from observed fMRI signals. In our earlier work~\cite{zhang2025mvho}, we introduced a matrix-based R\'enyi’s $\alpha$-order entropy functional estimator to enable reliable estimation without requiring explicit probability density function (PDF) modeling or data discretization. However, its $\mathcal{O}(N^3)$ computational complexity limited scalability, restricting the analysis to third-order HOIs.




In this work, we extend our previous work~\cite{zhang2025mvho} by moving beyond just three-way interactions and small-scale datasets, introducing scalable estimation strategies that enable higher-order (e.g., tetradic) modeling. Specifically, we develop two independent acceleration strategies: a closed-form analytical solution under Gaussian assumptions and a randomized approximation for the matrix-based R\'enyi’s $\alpha$-order entropy functional~\cite{yu2019multivariate}, both of which substantially reduce the computational cost of multivariate entropy and higher-order dependency estimation. These acceleration strategies make it feasible to extend HOI modeling from third to fourth order on benchmark cohorts, enabling a multi-view learning framework that jointly captures pairwise, triadic, and tetradic interactions while maintaining computational tractability.

In summary, our main contributions include:
\begin{itemize}
  \item We propose MvHo-IB++, a scalable multi-view information bottleneck framework that systematically models pairwise, third-order, and fourth-order brain interactions via $O$-information, while compressing redundant view-specific information with an information bottleneck-based regularization for fMRI-based psychiatric diagnosis.

  \item We develop two scalable strategies for higher-order $O$-information estimation. A Gaussian analytical approximation and a randomized matrix-based Rényi entropy approximation make third- and fourth-order interaction modeling computationally feasible on large-scale cohorts.
  
  \item We demonstrate the effectiveness and interpretability of the proposed framework on four benchmark datasets (REST-meta-MDD~\cite{restmeatmdd}, ABIDE~\cite{abide}, UCLA~\cite{ucla}, and ADNI~\cite{adni}). MvHo-IB++ achieves consistent improvements over state-of-the-art GNN and hypergraph-based baselines, while revealing disorder-related synergy- and redundancy-dominated higher-order interaction patterns that provide additional neuroimaging insights.
  
\end{itemize}

\section{Related Work}

\subsection{Multi-view Learning for fMRI-based Psychiatric Diagnosis}

Single-view approaches that rely solely on pairwise functional connectivity are limited to linear, dyadic dependencies and remain sensitive to noise and subject-specific artifacts~\cite{battiston2020networks, prado2022neurobiology, power2012spurious}. Multi-view learning addresses these challenges by integrating complementary representations of the same subject~\cite{jie2018integration}. Within the fMRI domain, multi-view pipelines have been designed along several dimensions: temporal views combine static and dynamic connectivity to capture both persistent and transient network patterns~\cite{allen2014tracking}; frequency-resolved views expose band-specific coupling mechanisms~\cite{wu2008frequency}; spatial views fuse multiple parcellations to reduce atlas dependence~\cite{glasser2016multi}; and interaction-order views, most relevant to this work, supplement pairwise connectivity with higher-order representations encoding concurrent interactions among three or more regions~\cite{battiston2020networks}.

Recent architectures instantiate these principles: multi-view hypergraph embeddings preserve higher-order relations across sites and atlases~\cite{CcSiMHAHGEL}; dual-branch networks fuse Euclidean connectivity features with non-Euclidean hypergraph representations~\cite{MHNet}; temporal--spatial models integrate dynamic and topological properties~\cite{jie2018integration}; and graph-based multi-view methods employ metric learning and spectral convolutions~\cite{ktena2018metric}. These approaches collectively underscore a shift from single-descriptor pipelines toward principled integration of complementary views, setting the stage for methods that systematically combine pairwise and higher-order interaction structures.

\subsection{Hypergraph-based modeling of higher-order brain interactions}
Hypergraphs extend pairwise graphs by allowing hyperedges that connect arbitrary-size sets of brain regions or networks, thereby enabling the representation of higher-order and multivariate functional interactions beyond dyadic connectivity. In fMRI, hyper-nodes correspond to anatomically segregated regions and hyperedges encode concurrent interactions among three or more regions, offering a more neurobiologically plausible representation of network-level dynamics than pairwise graphs. Recent brain-network studies have therefore adopted hypergraph neural operators to learn from such higher-order structures~\cite{wang2023dynamic}, and devised hyperedge-aware embeddings tailored to multisite, multi-atlas rs-fMRI.

Three principal strategies have emerged for constructing fMRI hypergraphs: learnable end-to-end approaches that optimize hyperedge topology via backpropagation~\cite{Hybrid}; shared representative structures that pre-compute a fixed topology from cohort-level statistics~\cite{HL}; and subject-specific structures derived from individual time series~\cite{CcSiMHAHGEL}. Despite these advances, hypergraph construction often relies on predefined similarity metrics and pruning heuristics~\cite{cai2022hypergraph}, and the resulting hyperedges typically encode multivariate co-association patterns without explicitly characterizing whether interactions are redundancy- or synergy-dominated~\cite{scagliarini2024gradients}.


\subsection{Information Bottleneck frameworks for brain network representation}\label{sec:related_3}

The information bottleneck (IB) principle~\cite{tishby2000information} seeks a representation $Z$ that is maximally predictive of the target $Y$ while compressing nuisance information about the input. For graph-structured data $G$ (e.g., fMRI-derived connectivity), the graph information bottleneck (GIB)~\cite{wu2020graph} learns an encoder $Z=f_{\phi}(G)$ by trading off sufficiency and compression:
\begin{equation}
\mathcal{L}_{\text{GIB}} = \min I(Z;G) \;-\; \beta\, I(Z;Y),
\end{equation}
with $\beta>0$. 
In practice, $I(Z;Y)$ is optimized via the cross-entropy surrogate based on the variational lower bound $I(Z;Y)\!\ge\! H(Y)+\mathbb{E}[\log P_{\theta}(Y\!\mid\!Z)]$, where $P_{\theta}(Y\mid Z)$ denotes the parametric predictive distribution (e.g., implemented by a classifier head) that maps the learned representation $Z$ to the target label $Y$.
$I(Z;G)$ is regulated through entropy regularization on $Z$. For deterministic encoders, $I(G;Z)=H(Z)$~\cite{strouse2017deterministic}, yielding the training objective:
\begin{equation}
\min_{\phi,\theta}\; \mathrm{CE}(Y,\hat Y)+\beta H(Z).
\end{equation}

When multiple views of the same subject are available (e.g., pairwise connectivity and higher-order interaction tensors), GIB extends to a multi-view setting:
\begin{equation}
\max I(Y;Z)\;-\;\big[\beta_1 I(X_1;Z_1)+\beta_2 I(X_2;Z_2)\big],\quad \text{s.t.}, \quad Z=f_{\theta}(Z_1,Z_2),
\end{equation}
with the practical surrogate $\min \; \mathrm{CE}(Y,\hat Y)+\beta_1 H(Z_1)+\beta_2 H(Z_2)$. 


The IB principle has recently been used for fMRI analysis. The BrainIB~\cite{brainib} addresses interpretability by extracting minimal predictive subgraphs: it learns to identify and retain only those nodes and edges that are maximally informative about the diagnostic label while discarding task-irrelevant structures, thereby compressing the brain graph into a compact, interpretable subgraph representation. 
BrainIB++~\cite{brainib++} further extends this framework by incorporating dynamic graph-attention mechanisms that refine connectivity graphs through learned edge reweighting or pruning under the IB objective, improving both node feature aggregation and graph-level representations for schizophrenia classification. 


\section{Methodology}

Consider a dataset of brain signal recordings $\{X^i, Y^i\}_{i=1}^N$, where each recording $X^i \in \mathbb{R}^{C \times T}$ represents the raw blood-oxygenation-level-dependent (BOLD) signal detected in fMRI for the $i$-th patient. Here, $C$ denotes the number of regions of interest (ROIs), e.g., $116$ for AAL atlas~\cite{tzourio2002automated} or $105$ for ICA-driven brain network template~\cite{iraji2023identifying}, and $T$ indicates the signal duration. We use subscripts to denote the channel index of the BOLD signal, where $X^j_i$ represents the $1$D signal from the $i$-th ($1\leq i\leq C$) brain region/network for the $j$-th ($1\leq j\leq N$) patient. Our objective is to develop a classifier that maps raw brain signal $X$ to its diagnostic label $Y$.

\begin{figure*}[!htbp]
\centering
\includegraphics[width=\textwidth]{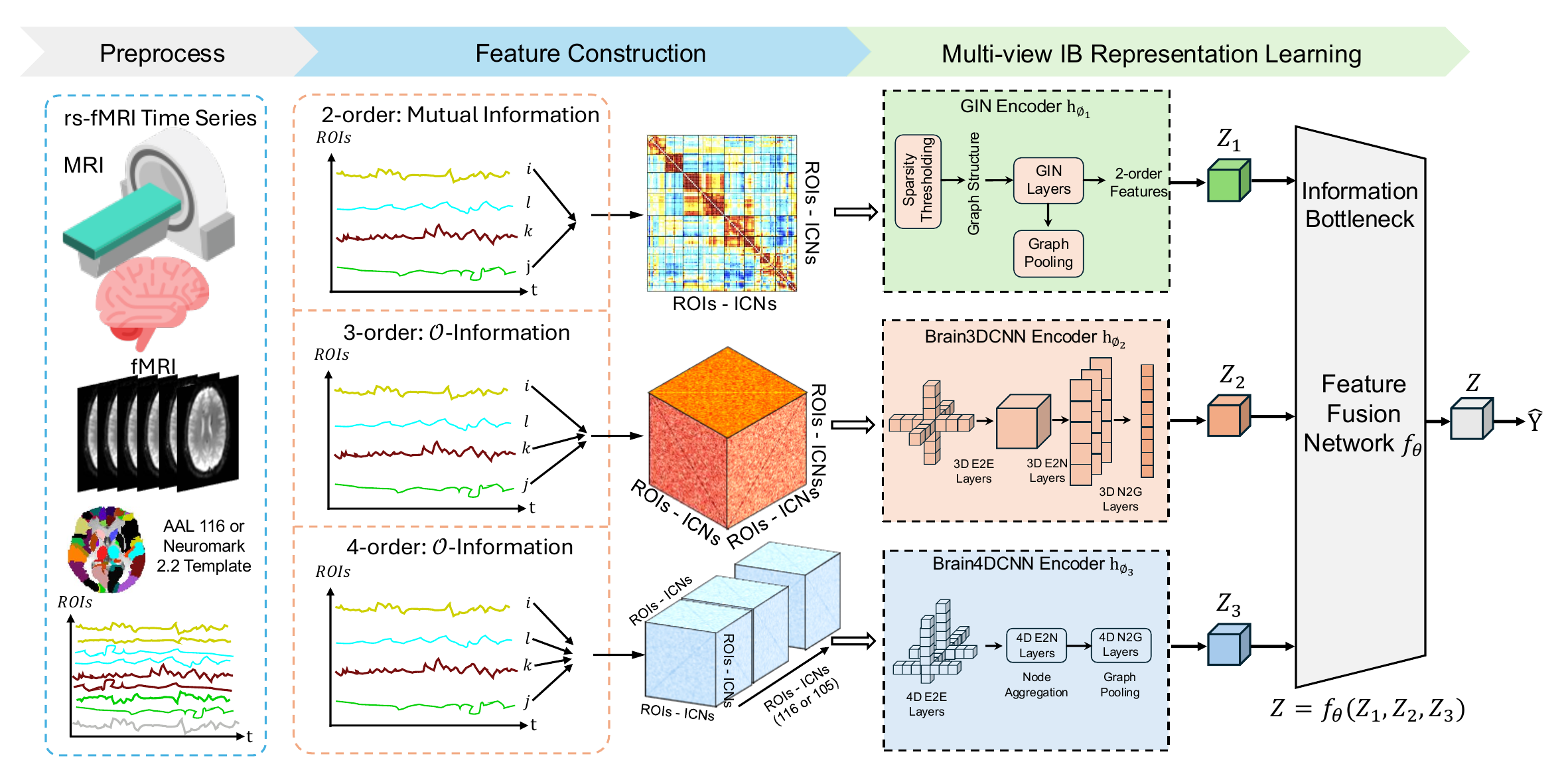}
\caption{Overview of the MvHo-IB++ framework. Raw rs-fMRI data are parcellated using the AAL-116 atlas or the Neuromark 2.2 template to extract ROI/ICN time series, from which three complementary views are constructed: a 2D pairwise mutual-information matrix, a 3D third-order $\mathcal{O}$-information tensor, and a 4D fourth-order $\mathcal{O}$-information tensor. Each view is processed by a specialized encoder (GIN $h_{\phi_1}$, Brain3DCNN $h_{\phi_2}$, and Brain4DCNN $h_{\phi_3}$) to produce latent representations $Z_1$, $Z_2$, and $Z_3$, respectively. These are integrated through a feature fusion network $f_\theta$ to obtain the unified representation $Z = f_\theta(Z_1, Z_2, Z_3)$, which is fed to a classifier to predict the diagnostic label $\hat{Y}$. The information bottleneck principle maximizes $I(Y;Z)$ while compressing view-specific redundancies.}
\label{fig:mvframework}
\end{figure*}

\subsection{Multi-view Information Bottleneck Framework}


We construct three complementary views from raw BOLD signals, as illustrated in Figure~\ref{fig:mvframework}. From ROI/ICN time series, we derive: (i) a $C \times C$ matrix encoding pairwise functional connectivity; (ii) a $C \times C \times C$ tensor capturing third-order interactions; and (iii) a $C \times C \times C \times C$ tensor representing fourth-order interactions. Together, these representations enable multi-order modeling of brain network dependencies, from pairwise correlations to higher-order interactions. 



The 2D view constructs a symmetric matrix $V_1 \in \mathbb{R}^{C \times C}$ where each element $V_1(i,j)$ represents the functional connectivity between brain regions $i$ and $j$. Instead of using the conventional Pearson's correlation coefficient, we employ matrix-based R\'enyi's $\alpha$-order mutual information~\cite{yu2019multivariate} to quantify nonlinear pairwise dependencies:
\begin{equation}
V_1(i,j) = I_{\alpha}(X_i; X_j)
\end{equation}
where $X_i$ and $X_j$ are BOLD time series from regions $i$ and $j$, respectively. To construct the graph structure for the GNN encoder, we apply a sparsity threshold retaining only the top 30\% of pairwise connections ranked by mutual information magnitude, setting the remaining entries to zero. 


The 3D view constructs a tensor $V_2 \in \mathbb{R}^{C \times C \times C}$ where each element $V_2(i,j,k)$ captures the three-way interaction among regions $i$, $j$, and $k$ using $\mathcal{O}$-information:
\begin{equation}
V_2(i,j,k) = \mathcal{O}(X_i, X_j, X_k).
\end{equation}


The 4D view extends to capture fourth-order interactions through a tensor $V_3 \in \mathbb{R}^{C \times C \times C \times C}$ where each element represents four-way interactions among quadruplets of brain regions/networks:
\begin{equation}
V_3(i,j,k,l) = \mathcal{O}(X_i, X_j, X_k, X_l)
\end{equation}



Figure~\ref{fig:mvframework} illustrates the architecture of MvHo-IB++. Specifically, our framework employs three specialized encoders: $h_{\phi_1}$ (Graph Isomorphism Network for the 2D connectivity matrix), $h_{\phi_2}$ (Brain3DCNN for the 3D third-order tensor), and $h_{\phi_3}$ (Brain4DCNN for the 4D fourth-order tensor), which produce latent representations $Z_1$, $Z_2$, and $Z_3$, respectively. These view-specific representations are integrated through a feature fusion network $f_\theta$ to obtain the unified representation $Z = f_\theta(Z_1, Z_2, Z_3)$, which is then fed to a classifier head $\omega$ to predict the diagnostic label $Y$.

The overall objective of the extended MvHo-IB++ framework is:
\begin{equation}
\begin{aligned}
&\mathop{\arg\max}_{\phi_1, \phi_2, \phi_3, \theta, w} \Bigl( I(Y; Z) - \Bigl( \beta_1 I(V_1; Z_1) + \beta_2 I(V_2; Z_2) + \beta_3 I(V_3; Z_3) \Bigr) \Bigr), \\
&\hspace{2.3cm} \text{s.t.} \quad Z = f_\theta(Z_1, Z_2, Z_3),
\end{aligned}
\label{eq:mvhoi_objective}
\end{equation}
where $\beta_1$, $\beta_2$, and $\beta_3$ are regularization coefficients for the three views, respectively.


As discussed in Section~\ref{sec:related_3}, for deterministic encoders, this objective reduces to:
\begin{equation}
\begin{aligned}
&\mathop{\arg\min}_{\phi_1, \phi_2, \phi_3, \theta, w} \Bigl( \operatorname{CE}(Y, \hat{Y}) + \beta_1 H(Z_1) + \beta_2 H(Z_2) + \beta_3 H(Z_3) \Bigr), \\
&\hspace{2.3cm} \text{s.t.} \quad Z = f_\theta(Z_1, Z_2, Z_3).
\end{aligned}
\label{eq:mvhoi_reformulated}
\end{equation}

The entropy terms $H(Z_1)$, $H(Z_2)$, and $H(Z_3)$ are estimated using the matrix-based R\'enyi's $\alpha$-order entropy functional (see Supplementary Material A for details). 

\subsection{$\mathcal{O}$-Information and Scalable Estimation}

We use $\mathcal{O}$-information to characterize HOIs among brain regions. For three BOLD time series $X_i, X_j, X_k \in \mathbb{R}^T$ corresponding to regions $i$, $j$, and $k$, the $\mathcal{O}$-information is defined as the difference between total correlation (TC) and dual total correlation (DTC):
\begin{equation}
\mathcal{O}(X_i,X_j,X_k)=\mathrm{TC}(X_i,X_j,X_k)-\mathrm{DTC}(X_i,X_j,X_k).
\end{equation}

Here, TC and DTC are multivariate dependence measures in information theory:
\begin{align}
\mathrm{TC}(X_i,X_j,X_k)&=H(X_i)+H(X_j)+H(X_k)-H(X_i,X_j,X_k), \label{eq:tc}\\
\mathrm{DTC}(X_i,X_j,X_k)&=H(X_i,X_j,X_k)-H(X_i\!\mid\!X_j,X_k)-H(X_j\!\mid\!X_i,X_k)\notag\\
&\qquad -H(X_k\!\mid\!X_i,X_j). \label{eq:dtc}
\end{align}

\begin{figure}[!htbp]
\centering
\includegraphics[width=0.48\textwidth]{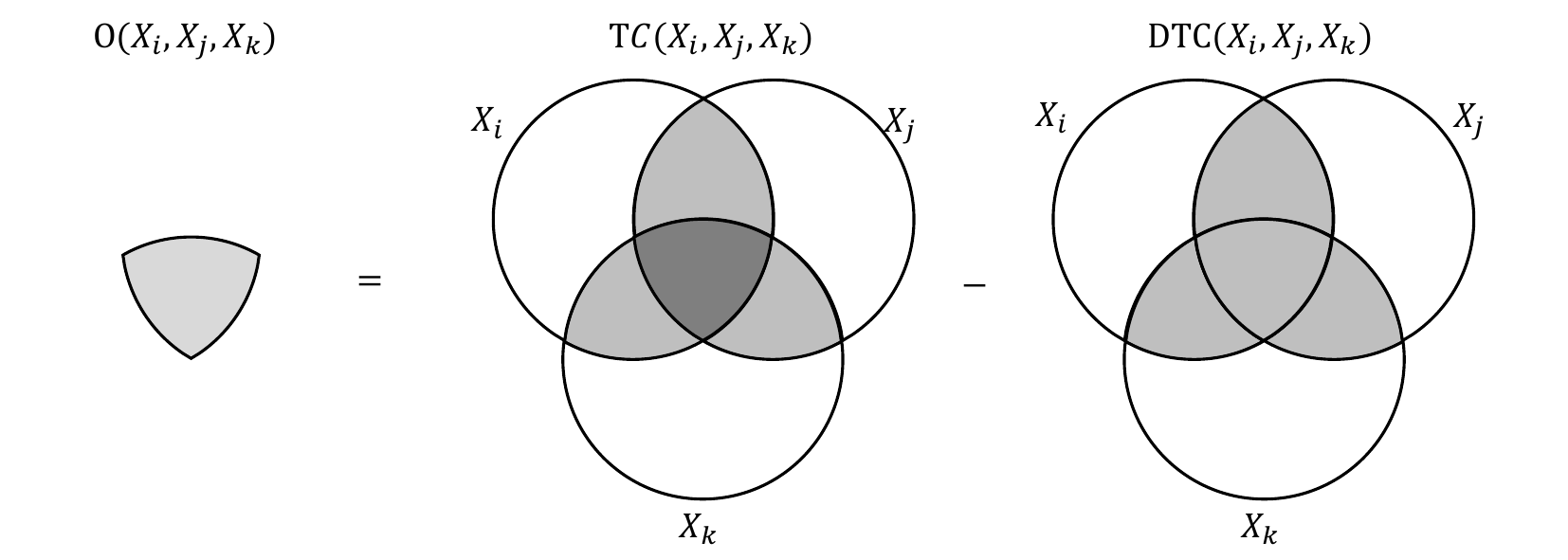}
\caption{Venn diagram illustration of total correlation (TC) and dual total correlation (DTC).}
\label{fig:venn}
\end{figure}

From an information-theoretic perspective, a positive value of $\mathcal{O}$-information indicates that redundancy outweighs synergy among brain regions $i$, $j$, and $k$, meaning that their joint interactions contain predominantly redundant information. In contrast, a negative value reflects synergy-dominated interactions~\cite{rosas2019oinfo,scagliarini2024gradients}. A Venn diagram illustrating the relationships among TC, DTC, and $\mathcal{O}$-information is shown in Figure~\ref{fig:venn}.

The fourth-order $\mathcal{O}$-information is defined analogously as:
\begin{equation}
\mathcal{O}(X_i,X_j,X_k,X_\ell)=\mathrm{TC}(X_i,X_j,X_k,X_\ell)-\mathrm{DTC}(X_i,X_j,X_k,X_\ell).
\end{equation}

Direct estimation of $\mathcal{O}$-information at scale is computationally expensive. We therefore adopt two alternative acceleration strategies, used independently: 1) a Gaussian analytical approximation, and 2) a randomized approximation for matrix-based Rényi entropy. The Gaussian approximation provides a closed-form and efficient solution under second-order statistics, while the randomized Rényi estimator offers a more general nonparametric alternative.
Both strategies exploit the symmetry of covariance and Gram matrices by operating only on their upper-triangular entries, thereby avoiding redundant computations and reducing memory consumption.

\subsubsection{Gaussian analytical approximation}
Under the multivariate Gaussian assumption, consider a $K$-tuple of brain regions $X_1,\dots,X_K$. The TC among them can be computed as:
\begin{equation}
\widehat{\mathrm{TC}}(X_1,\dots,X_K)
=
\frac{1}{2}\log
\left(
\frac{\det(\mathrm{diag}(\Sigma))}
{\det(\Sigma)}
\right),
\end{equation}
where $\Sigma \in \mathbb{R}^{K\times K}$ is the covariance matrix of
$(X_1,\dots,X_K)$.

The DTC also admits a closed-form log-determinant expression based on principal minors of $\Sigma$. The $\mathcal{O}$-information is then given by:
\begin{equation}
\mathcal{O}(X_1,\dots,X_K)
= \widehat{\mathrm{TC}}(X_1,\dots,X_K)
- \widehat{\mathrm{DTC}}(X_1,\dots,X_K).
\end{equation}

This formulation requires only covariance determinants and principal minors, avoiding density estimation. For small fixed $K$ (e.g., $K=3,4$ in our work), the total computational complexity scales approximately as $\mathcal{O}(C^K T)$ over all region combinations. Detailed derivations and complexity analysis are provided in Supplementary Material B.

\subsubsection{Randomized approximation for matrix-based Rényi's entropy.}

As an alternative that does not rely on distributional assumptions, we estimate entropies using the matrix-based R\'enyi entropy functional of order $\alpha$. Given $T$ time-point samples and the normalized Gram matrix $\mathbf{G} \in \mathbb{R}^{T \times T}$ constructed with a Gaussian kernel, the entropy is defined as
\begin{equation}
H_\alpha(X)
= \frac{1}{1-\alpha}
\log \big( \mathrm{tr}(\mathbf{G}^\alpha) \big)
= \frac{1}{1-\alpha}
\log \left(\sum_{i=1}^{T}\lambda_i^\alpha \right),
\end{equation}
$\{\lambda_i\}_{i=1}^T$ denote the eigenvalues of $\mathbf{G}$.

To avoid explicit eigendecomposition with cubic complexity $\mathcal{O}(T^3)$, we approximate the trace using randomized trace estimation~\cite{random}, yielding an efficient estimator $\widetilde{H}_\alpha(X)$. Specifically, let $\{\mathbf{g}_i\}_{i=1}^s$ be i.i.d. standard Gaussian probe vectors. Then
\begin{equation}
\widetilde{H}_{\alpha}(X)
=
\frac{1}{1-\alpha}
\log\!\left(
\frac{1}{s}
\sum_{i=1}^{s}
\mathbf{g}_i^{\top}
\mathbf{G}^{\alpha}
\mathbf{g}_i
\right).
\end{equation}

Using this entropy approximation estimator, we can efficiently evaluate each entropy term in both TC and DTC (see Eqs.~\ref{eq:tc} and~\ref{eq:dtc}); consequently, the O-information can be computed as follows:
\begin{equation}
\mathcal{O}_\alpha = TC_\alpha - DTC_\alpha.
\end{equation}

For fixed $\alpha$ and a small number of probe vectors, the overall complexity scales approximately as $\mathcal{O}(C^K T^2)$ over all $K$-tuples. Detailed derivations and algorithmic steps are provided in Supplementary Material B.

\subsection{BrainCNN encoders for high-order interactions}

To process the third-order and fourth-order $\mathcal{O}$-information tensors, we develop Brain3DCNN and Brain4DCNN architectures with specialized cross-shaped convolution kernels. Both follow a hierarchical edge-to-node-to-graph processing paradigm. We detail Brain4DCNN below, as Brain3DCNN has been described in our prior work~\cite{zhang2025mvho}.

\begin{figure}[!htbp]
\centering
\includegraphics[width=0.5\textwidth]{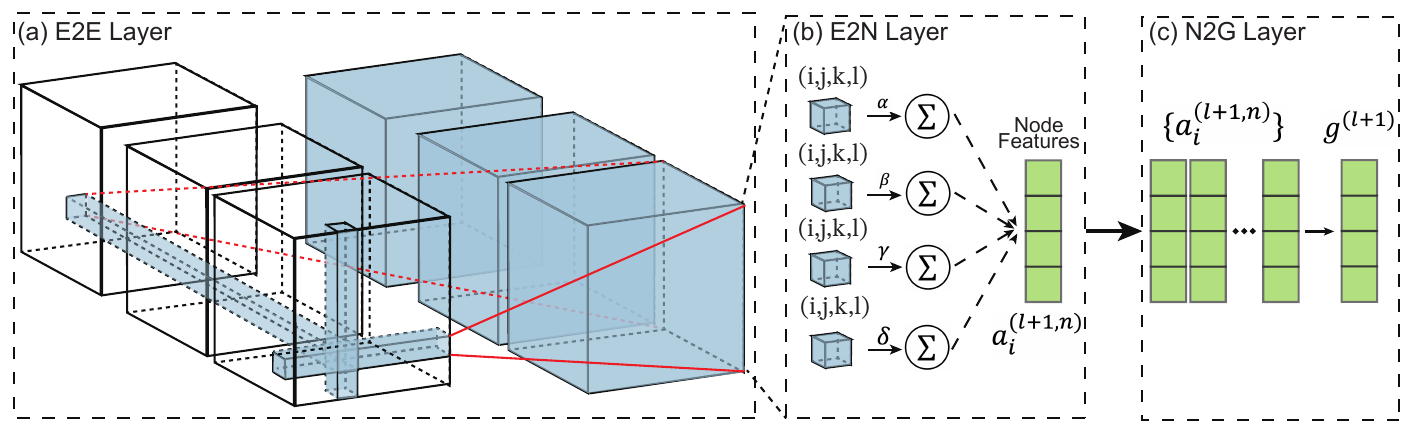}
\caption{Architecture of Brain4DCNN with three hierarchical layers. (a) The 4D Edge-to-Edge (E2E) layer applies four-dimensional cross-shaped convolution kernels to the input fourth-order $\mathcal{O}$-information tensor, capturing local connectivity patterns along all four spatial axes simultaneously. (b) The 4D Edge-to-Node (E2N) layer aggregates fourth-order hyperedge representations $(i,j,k,l)$ into node-level features through weighted summation with learnable parameters $\alpha$, $\beta$, $\gamma$, and $\delta$ that capture position-specific contributions. (c) The 4D Node-to-Graph (N2G) layer performs global pooling over all node features $\{a_i^{(\ell+1,n)}\}$ to produce the graph-level representation $g^{(\ell+1)}$ for classification.}
\label{fig:brain4dcnn}
\end{figure}

Figure~\ref{fig:brain4dcnn} illustrates the three-stage hierarchical architecture of Brain4DCNN for processing fourth-order brain interactions.

\subsubsection{4D Edge-to-Edge (E2E) Layer}


To capture fourth-order interactions encoded in the $\mathcal{O}$-information tensor, we extend the 2D Edge-to-Edge (E2E) architecture~\cite{kawahara2017brainnetcnn} to a 4D E2E layer.
As shown in Figure~\ref{fig:brain4dcnn}(a), the layer processes
a fourth-order interaction tensor using a structured
cross-shaped convolutional kernel.


Formally, let $\mathbf{O}^{(\ell)} \in \mathbb{R}^{M^\ell \times C \times C \times C \times C}$ denote all feature maps at the $\ell$-th layer, extracted from $M^\ell$ convolutional kernels. For the first input layer, $\mathbf{O}^{1} \in \mathbb{R}^{C \times C \times C \times C}$ is the estimated fourth-order $\mathcal{O}$-information tensor.

We adopt a cross-shaped kernel that performs one-dimensional convolutions independently along each of the four spatial axes,
followed by a linear combination.
For the $n$-th output feature map at layer $(\ell+1)$, the response is computed as:
\begin{equation}
\begin{aligned}
O^{(\ell+1,n)}_{i,j,k,l} = &\sum_{m=1}^{M^\ell} \sum_{c=1}^{K} \Bigl[
w^{(\ell,m,n)}_{1,c}\,O^{(\ell,m)}_{i+c,\,j,\,k,\,l} + w^{(\ell,m,n)}_{1,-c}\,O^{(\ell,m)}_{i-c,\,j,\,k,\,l} \\
&+ w^{(\ell,m,n)}_{2,c}\,O^{(\ell,m)}_{i,\,j+c,\,k,\,l} + w^{(\ell,m,n)}_{2,-c}\,O^{(\ell,m)}_{i,\,j-c,\,k,\,l} \\
&+ w^{(\ell,m,n)}_{3,c}\,O^{(\ell,m)}_{i,\,j,\,k+c,\,l} + w^{(\ell,m,n)}_{3,-c}\,O^{(\ell,m)}_{i,\,j,\,k-c,\,l} \\
&+ w^{(\ell,m,n)}_{4,c}\,O^{(\ell,m)}_{i,\,j,\,k,\,l+c} + w^{(\ell,m,n)}_{4,-c}\,O^{(\ell,m)}_{i,\,j,\,k,\,l-c} \Bigr]
\end{aligned}
\label{eq:e2e_4d}
\end{equation}
where $w^{(\ell,m,n)}_{d,\pm c}$ are learnable weights for the cross-shaped kernel in dimension $d \in \{1,2,3,4\}$, and $K$ is the kernel radius. 

This design preserves the structural symmetry of fourth-order interactions while significantly reducing the number of parameters from $\mathcal{O}(K^4)$ to $\mathcal{O}(4K)$ per kernel.


\subsubsection{4D Edge-to-Node (E2N) Layer}

As illustrated in Figure~\ref{fig:brain4dcnn}(b), the 4D E2N layer aggregates the fourth-order interaction tensor into node-level representations while preserving higher-order dependence structure. 

Given $\mathbf{O}^{(\ell)} \in \mathbb{R}^{M^\ell \times C \times C \times C \times C}$ from the 4D E2E layer, the E2N layer computes the representation of region $i$ by aggregating all fourth-order interactions involving $i$.


Specifically, for the $n$-th output channel,
\begin{equation}
\begin{aligned}
a_i^{(\ell+1,n)} = \sum_{m=1}^{M^\ell} \sum_{j,k,l=1}^{C} \Bigl[
&\alpha^{(\ell,m,n)}\,O^{(\ell,m)}_{i,j,k,l} + \beta^{(\ell,m,n)}\,O^{(\ell,m)}_{j,i,k,l} \\
&+ \gamma^{(\ell,m,n)}\,O^{(\ell,m)}_{j,k,i,l} + \delta^{(\ell,m,n)}\,O^{(\ell,m)}_{j,k,l,i} \Bigr].
\end{aligned}
\label{eq:e2n_4d}
\end{equation}


The learnable weights $\alpha^{(\ell,m,n)}$, $\beta^{(\ell,m,n)}$, $\gamma^{(\ell,m,n)}$, and $\delta^{(\ell,m,n)}$ capture the contribution of region $i$ when it appears in different tensor modes. This aggregation preserves permutation symmetry across modes, ensuring that all four-way interactions involving region $i$ are integrated into its node representation.

\subsubsection{4D Node-to-Graph (N2G) Layer}

As depicted in Figure~\ref{fig:brain4dcnn}(c), the 4D N2G layer performs the final aggregation of node-level representations into a graph-level feature vector that summarizes fourth-order interaction patterns across the entire brain network. 


Let $\{a_i^{(\ell+1)}\}_{i=1}^{C}$ denote the node representations produced by the 4D E2N layer. The graph-level feature is obtained via permutation-invariant pooling:
\begin{equation}
g^{(\ell+1)} = \mathrm{Pool}\left(\{a_i^{(\ell+1,n)} : i = 1, \ldots,C\}\right).
\end{equation}

In our implementation, we employ global average pooling, though other permutation-invariant operators (e.g., max pooling or attention-based aggregation) can be used. This operation summarizes higher-order interaction information into a compact graph-level representation
for downstream classification.




\section{Experiments}
\subsection{Datasets and Experimental Settings}
We benchmark our approach on four publicly available rs-fMRI cohorts. 
(i) The UCLA Consortium for Neuropsychiatric Phenomics dataset~\cite{ucla} comprises schizophrenia (SZ; $n{=}50$) and normal control (NC; $n{=}114$) participants. 
(ii) The Alzheimer's Disease Neuroimaging Initiative (ADNI)~\cite{adni} contains subjects with mild cognitive impairment (MCI; $n{=}38$) and matched NCs ($n{=}37$). 
(iii) The Autism Brain Imaging Data Exchange I (ABIDE-I)~\cite{abide} aggregates multi-site rs-fMRI from more than 1{,}000 individuals.\footnote{\url{http://fcon_1000.projects.nitrc.org/indi/abide/}} 
In our study, we use 528 autism spectrum disorder (ASD) patients and 571 typically developing (TD) controls. 
(iv) The REST-meta-MDD consortium~\cite{restmeatmdd} is the largest MDD rs-fMRI resource to date, originally including 2{,}428 scans (1{,}300 MDD and 1{,}128 healthy controls, HCs) collected across 25 research groups in 17 hospitals in China. 
Following standard exclusion criteria (incomplete metadata, suboptimal normalization or coverage, excessive head motion, sites with fewer than ten subjects per group, and incomplete time series), 1{,}604 scans remained for analysis (828 MDD, 776 HCs).
Demographic and clinical summaries for all four datasets appear in Table~\ref{tab:demographics}.

For UCLA, we applied group-level independent component analysis using the NeuroMark\_fMRI\_2.2 template\footnote{\url{https://trendscenter.org/data/}} to derive 105 spatially independent functional networks from the consortium-provided preprocessed data~\cite{iraji2023identifying}.
For ABIDE, we performed a conventional SPM-based pipeline\footnote{\url{https://www.fil.ion.ucl.ac.uk/spm/software/spm12/}} consisting of slice-timing correction, rigid-body realignment, spatial normalization to MNI space, spatial smoothing with a 6\,mm FWHM Gaussian kernel, band-pass filtering (0.01--0.08\,Hz), and regression of nuisance signals (head motion parameters, white matter, and cerebrospinal fluid).
For REST-meta-MDD, we used the consortium-provided preprocessed time series, generated via a harmonized workflow: discard the first ten volumes, slice-timing correction, motion correction, normalization to MNI space, temporal filtering (0.01--0.1\,Hz), and regression of confounds including motion parameters, global signal, white matter, cerebrospinal fluid, and linear trends. Note that the different filtering parameters across datasets reflect adherence to each consortium's standardized preprocessing protocol.
After preprocessing, we extracted time series from functional networks or regions using spatially constrained ICA with the NeuroMark 2.2 multiscale template~\cite{jensen2024addressing}, available at http://trendscenter.org/data and in the GIFT toolbox (http://trendscenter.org/software/gift): for UCLA, 105 ICA component time courses were obtained; for ABIDE, ADNI, and REST-meta-MDD, we extracted mean BOLD time series from 116 regions of interest (ROIs) (AAL atlas: 90 cerebral and 26 cerebellar regions). 


\begin{table*}[h]
\centering
\caption{Demographic and clinical characteristics for all datasets.}
\label{tab:demographics}
\resizebox{\textwidth}{!}{%
\begin{tabular}{lcccccccc}
\hline
\multirow{2}{*}{\textbf{Characteristic}} & \multicolumn{2}{c}{\textbf{UCLA}} & \multicolumn{2}{c}{\textbf{ABIDE}} & \multicolumn{2}{c}{\textbf{ADNI}} & \multicolumn{2}{c}{\textbf{REST-meta-MDD}} \\
\cline{2-9}
 & SZ & HC & ASD & TD & MCI & NC & MDD & HC \\
\hline
Sample size & 50 & 114 & 528 & 571 & 38 & 37 & 828 & 776 \\
Age (years) & $39.17 \pm 8.59$ & $31.20 \pm 8.76$ & $17.0 \pm 8.4$ & $17.1 \pm 7.7$ & $73.1 \pm 7.3$ & $74.1 \pm 6.2$ & $34.3 \pm 11.5$ & $34.4 \pm 13.0$ \\
Age range & 21--50 & 21--50 & 7--64 & 8.1--56.2 & 60.0--88.7 & 65.2--90.0 & 18--65 & 18--64 \\
Gender (M/F) & 42/8 & 58/56 & 464/64 & 471/100 & 19/19 & 15/22 & 301/527 & 318/458 \\
Education (years) & -- & -- & -- & -- & $15.6 \pm 2.9$ & $16.5 \pm 2.3$ & $12.0 \pm 3.4$ & $13.6 \pm 3.4$ \\
\hline
\end{tabular}%
}
\end{table*}

\subsection{Baselines and Evaluation Protocol}

We compare against eleven representative methods that cover prevailing design families for brain network modeling: three foundational graph neural networks, GCN~\cite{kipf2016semi}, GAT~\cite{velickovic2017graph}, and GIN~\cite{xu2018gin}; three information-theoretic models, SIB~\cite{SIB}, DIRGNN~\cite{dirgnn}, and BrainIB~\cite{brainib}; a multi-view model operating on standard graph and Euclidean space, MHNet~\cite{MHNet}; and three neuroscience-informed hypergraph construction pipelines, HyBRiD~\cite{Hybrid}, HL~\cite{HL}, and CcSi\mbox{-}MHAHGEL~\cite{CcSiMHAHGEL}. We also compare with the original MvHo-IB~\cite{zhang2025mvho}, which is limited to two views and incurs substantial computational cost.

As HL and CcSi\mbox{-}MHAHGEL are not open-source, we re-implemented their hypergraph construction procedures following the original descriptions; all hypergraph baselines share a common HyperAtten~\cite{gao2022hgnn+} backbone for fair comparison (implementation details are provided in the Supplementary Material, Section~D).

\subsubsection{Evaluation Protocol}

We adopt two complementary validation schemes to quantify generalization. First, we perform tenfold cross validation within each dataset and report the mean and standard deviation across folds. Second, on the multi-site cohort REST-meta-MDD, we conduct leave-one-site-out cross validation to assess out-of-site performance on unseen acquisition centers. All methods use matched preprocessing, a common atlas, and identical graph or hypergraph construction settings. Hyperparameters follow authors' recommendations and are tuned within the same search ranges. Model selection is carried out on validation splits in each fold or held-out site, and test metrics are reported once per split with identical evaluation criteria across all competitors.

\subsection{Hyperparameter Setup}
All models were implemented in PyTorch and trained on a single NVIDIA A100 40\,GB GPU.
We used Adam with an initial learning rate of \(1{\times}10^{-5}\), halved every 50 epochs, and weight decay of 0.03.
The GIN backbone contains three graph layers with two-layer MLP blocks of hidden sizes [128, 256, 512], followed by batch normalization and ReLU.
The Brain3DCNN branch comprises E2E3D layers (32 to 64 channels) with spatial–channel reduction.
Training ran for 100 epochs with batch size 32 and dropout 0.5.
For the matrix-based Rényi's $\alpha$-order entropy functional used in the information bottleneck loss (Eq.~\ref{eq:mvhoi_reformulated}), we set the Gaussian kernel bandwidth $\sigma{=}5$ and $\alpha{=}1.01$. For its randomized approximation, the number of $s$ is 30 in all experiments.
Regularization coefficients $\beta_1,\beta_2,\beta_3 \in \{0,\,10^{-4},\,10^{-3},\,10^{-2},\,10^{-1}\}$ were selected via tenfold cross-validation, with the best validation accuracy obtained at $\beta_1{=}0.01$, $\beta_2{=}0.1$, and $\beta_3{=}0.1$.
The fusion module aggregates the three learned views using a three-layer MLP with ReLU and dropout ($p{=}0.5$).
Competing baselines were configured according to their published recommendations and tuned within their prescribed ranges.
\section{Results}

\subsection{Overall comparison}
Table~\ref{tab:comparison} shows that \textsc{MvHo-IB++} achieves the best accuracy on all four datasets. The gains over the strongest competing methods range from 0.94\% to 3.98\%, demonstrating consistent improvements across cohort sizes and diagnostic categories. The performance of the original \textsc{MvHo-IB} on REST-meta-MDD and ABIDE is omitted, as the method does not scale to large datasets due to its high computational complexity.

\begin{table}[h]
\centering
\caption{Classification accuracy on four rs-fMRI datasets. Best results are in \textbf{bold}.}
\label{tab:comparison}
\resizebox{0.5\textwidth}{!}{%
\begin{tabular}{lcccc}
\hline
\textbf{Method} & \textbf{UCLA} & \textbf{ADNI} & \textbf{REST} & \textbf{ABIDE} \\
\hline
GCN~\cite{kipf2016semi} & 62.27 $\pm$ 6.21 & 66.13 $\pm$ 4.62 & 60.82 $\pm$ 3.87 & 64.48 $\pm$ 3.94 \\
GAT~\cite{velickovic2017graph} & 67.73 $\pm$ 7.61 & 66.28 $\pm$ 8.69 & 64.77 $\pm$ 4.13 & 67.93 $\pm$ 4.27 \\
GIN~\cite{xu2018gin} & 65.91 $\pm$ 8.21 & 68.33 $\pm$ 6.47 & 65.42 $\pm$ 3.96 & 67.95 $\pm$ 4.18 \\
\hline
DIR-GNN~\cite{dirgnn} & 75.72 $\pm$ 8.37 & 70.63 $\pm$ 6.96 & 64.53 $\pm$ 4.21 & 72.07 $\pm$ 3.84 \\
SIB~\cite{SIB} & 72.76 $\pm$ 8.13 & 70.21 $\pm$ 7.43 & 55.72 $\pm$ 5.28 & 62.74 $\pm$ 4.67 \\
BrainIB~\cite{brainib} & 79.14 $\pm$ 4.17 & 72.47 $\pm$ 5.32 & 70.02 $\pm$ 3.21 & 70.22 $\pm$ 3.58 \\
\hline
MHNet~\cite{MHNet} & 79.22 $\pm$ 6.72 & 71.96 $\pm$ 4.96 & 71.03 $\pm$ 3.14 & 71.49 $\pm$ 3.42 \\
HyBRiD~\cite{Hybrid} & 79.38 $\pm$ 8.34 & 71.34 $\pm$ 7.43 & 70.86 $\pm$ 3.27 & 71.72 $\pm$ 3.51 \\
HL~\cite{HL} & 78.74 $\pm$ 7.21 & 70.61 $\pm$ 6.84 & 70.16 $\pm$ 3.36 & 70.05 $\pm$ 3.62 \\
CcSi\mbox{-}MHAHGEL~\cite{CcSiMHAHGEL} & 76.84 $\pm$ 7.86 & 70.18 $\pm$ 7.21 & 68.77 $\pm$ 3.74 & 69.01 $\pm$ 3.89 \\
\hline
MvHo-IB~\cite{zhang2025mvho} & 83.12 $\pm$ 5.74 & 73.23 $\pm$ 4.37 & -- & -- \\
\hline
\textbf{MvHo-IB++} & \textbf{83.36 $\pm$ 5.63} & \textbf{73.41 $\pm$ 4.26} & \textbf{72.53 $\pm$ 2.54} & \textbf{72.61 $\pm$ 2.68} \\
\hline
\end{tabular}%
}
\end{table}

Conventional GNN baselines (GCN, GAT, and GIN) operate on pairwise functional connectivity graphs and consistently underperform \textsc{MvHo-IB++} across cohorts. 

Information-theoretic baselines (DIR-GNN, SIB, and BrainIB) improve upon vanilla GNNs by regularizing representations to preserve task-relevant information while suppressing noise. However, these methods remain constrained by single-view pairwise graphs and therefore cannot fully exploit complementary signals across interaction orders. In particular, approaches based on subgraph selection or pruning can be sensitive to cohort heterogeneity, which may degrade robustness when acquisition protocols vary. 

Hypergraph and multi-view baselines (MHNet, HyBRiD and the HyperAtten variants HyperAtten$_{\text{HL}}$ and HyperAtten$_{\text{CcSi\mbox{-}MHAHGEL}}$) partially capture higher-order structure via hyperedge construction or attention-based aggregation, and they perform competitively on several cohorts. Nonetheless, their higher-order modeling is either implicit or dependent on heuristic hypergraph design choices, which can miss interaction patterns that are not well represented by constructed hyperedges. \textsc{MvHo-IB++} instead provides a systematic, order-explicit characterization through multi-order $\mathcal{O}$-information tensors, and the specialized Brain4DCNN encoder directly learns from these tensors.

\subsection{Ablation Study}

To disentangle the contribution of each component in \textsc{MvHo-IB++}, we conduct an ablation study on all four datasets; Table~\ref{tab:ablation} summarizes the results.

\begin{table}[h]
\centering
\caption{Ablation study on four rs-fMRI datasets. Accuracy (\%) is reported.}
\label{tab:ablation}
\resizebox{0.48\textwidth}{!}{%
\begin{tabular}{lcccc}
\hline
\textbf{Configuration} & \textbf{UCLA} & \textbf{ADNI} & \textbf{REST} & \textbf{ABIDE} \\
\hline
GIN (MI) & 66.17 $\pm$ 7.84 & 68.74 $\pm$ 6.21 & 64.75 $\pm$ 3.92 & 67.02 $\pm$ 4.13 \\
Brain3DCNN (Gauss) & 73.96 $\pm$ 6.47 & 70.17 $\pm$ 5.68 & 66.61 $\pm$ 3.51 & 70.17 $\pm$ 3.76 \\
Brain3DCNN (Random) & 74.09 $\pm$ 6.38 & 70.31 $\pm$ 5.54 & 66.68 $\pm$ 3.48 & 70.32 $\pm$ 3.71 \\
Brain4DCNN (Gauss) & 74.18 $\pm$ 6.29 & 70.41 $\pm$ 5.47 & 66.37 $\pm$ 3.53 & 70.39 $\pm$ 3.69 \\
2-view w/o IB & 81.77 $\pm$ 5.82 & 72.06 $\pm$ 4.63 & 71.21 $\pm$ 2.87 & 71.67 $\pm$ 3.14 \\
3-view w/o IB & 82.86 $\pm$ 5.71 & 72.94 $\pm$ 4.41 & 71.72 $\pm$ 2.73 & 72.18 $\pm$ 2.94 \\
\hline
\textbf{Full model} & \textbf{83.36 $\pm$ 5.63} & \textbf{73.41 $\pm$ 4.26} & \textbf{72.53 $\pm$ 2.54} & \textbf{72.61 $\pm$ 2.68} \\
\hline
\end{tabular}%
}
\end{table}

As shown in Table~\ref{tab:ablation}, \textit{GIN (MI)} uses only the second-order view constructed from matrix-based mutual information with a GIN encoder, which suggests that pairwise connectivity alone is insufficient for robust psychiatric classification. Moving to higher-order modeling, \textit{Brain3DCNN (Gauss)} encodes a third-order $\mathcal{O}$-information tensor computed via a Gaussian analytical approximation, while \textit{Brain3DCNN (Random)} uses a randomized approximation for matrix-based R\'enyi entropy. The two estimators yield near-identical performance (within 0.15 points), indicating accurate and stable entropy estimation for constructing $\mathcal{O}$-information tensors and supporting our dual-strategy design for efficiency. Extending to fourth order, \textit{Brain4DCNN (Gauss)} indicates that fourth-order interactions provide consistent but moderate improvements when used alone.

Multi-view integration substantially boosts performance. The \textit{2-view w/o IB} model fuses the second-order MI-based pairwise graph with the third-order (random-estimator) $\mathcal{O}$-information tensor without the information bottleneck objective, while \textit{3-view w/o IB} further adds the fourth-order view; both configurations outperform single-view higher-order models in Table~\ref{tab:ablation}. The improvement from 2-view to 3-view indicates that the fourth-order tensor provides complementary patterns that are not fully redundant with lower-order views, and that fusion is essential to exploit distinct signals at each order.

Applying the multi-view information bottleneck regularization to the 3-view architecture yields the \textit{full model}, which achieves the best accuracy on all four datasets in Table~\ref{tab:ablation}. This shows that the entropy regularization terms $\beta_1 H(Z_1) + \beta_2 H(Z_2) + \beta_3 H(Z_3)$ effectively compress view-specific redundancies and improve generalization. The information bottleneck principle encourages $Z_1, Z_2, Z_3$ to retain task-relevant information while discarding noise and spurious correlations, which is particularly beneficial on heterogeneous multi-site cohorts such as REST and ABIDE.

Overall, the ablations show that higher-order modeling provides the largest jump over pairwise baselines, multi-view fusion accounts for the main additional gains, the fourth-order view yields consistent but smaller improvements, and the information bottleneck objective adds modest yet reliable refinements. Together, these components enable \textsc{MvHo-IB++} to achieve state-of-the-art performance across all four datasets.

\subsection{Hyperparameter Sensitivity Analysis}

\begin{figure}[!htbp]
\centering
\includegraphics[width=0.48\textwidth]{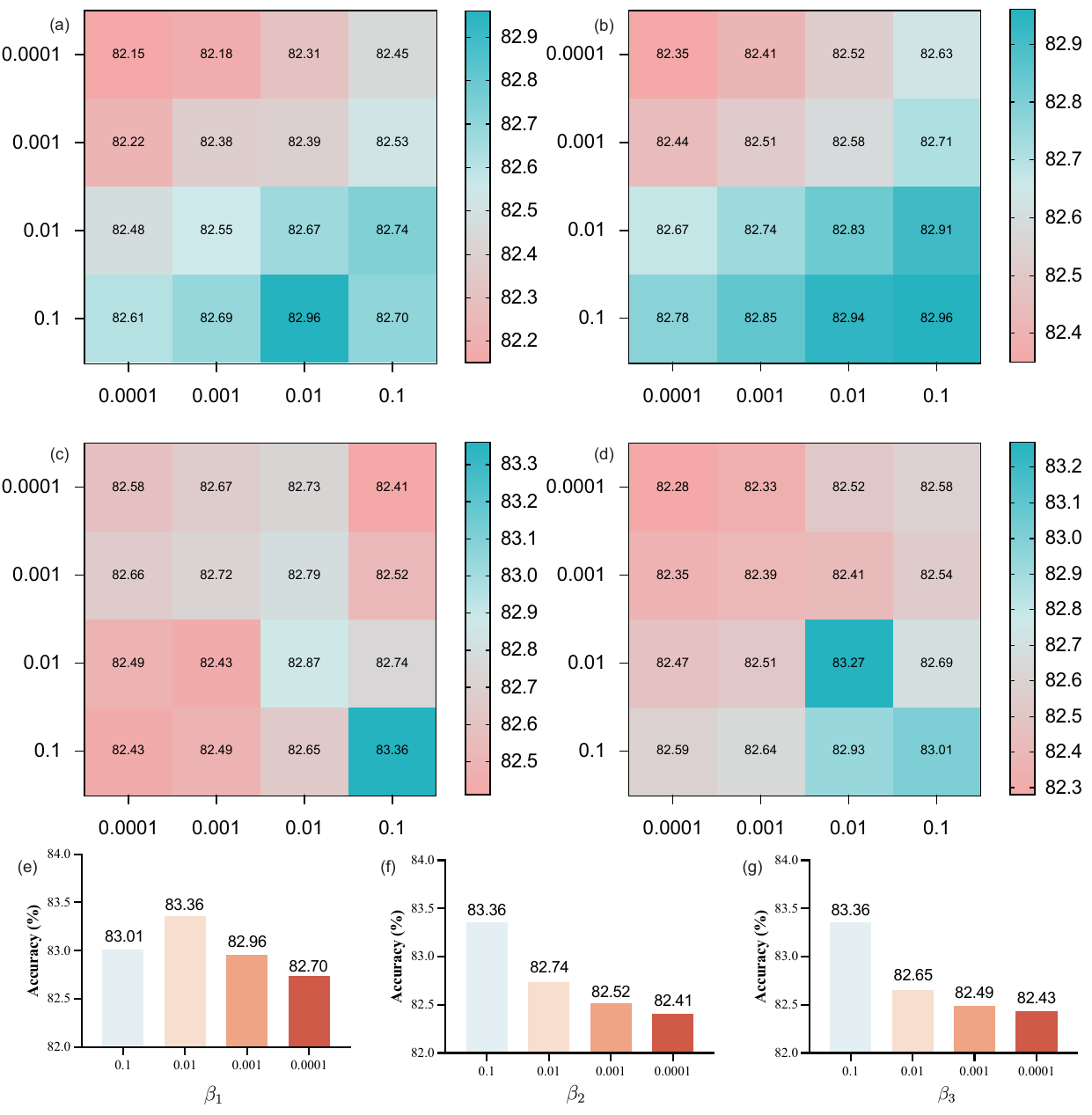}
\caption{Hyperparameter sensitivity analysis on the UCLA dataset. Panels (a), (b), (c), and (d) show classification accuracy heatmaps as a function of $\beta_2$ (x-axis) and $\beta_3$ (y-axis) for fixed $\beta_1$ values of 0.0001, 0.001, 0.01, and 0.1, respectively. Color intensity indicates accuracy, with warmer colors representing higher performance. Panels (e), (f), and (g) show individual sensitivity curves of $\beta_1$, $\beta_2$, and $\beta_3$ when the other two coefficients are fixed at their optimal values ($\beta_1{=}0.01$, $\beta_2{=}0.1$, $\beta_3{=}0.1$).}
\label{fig:heatmap}
\end{figure}

\textsc{MvHo-IB++} introduces three information bottleneck coefficients, $\beta_1$, $\beta_2$, and $\beta_3$, which control view-specific compression for the second-, third-, and fourth-order representations, respectively, and a Gaussian kernel bandwidth $\sigma$ in the matrix-based R\'enyi entropy estimator. We assess robustness by sweeping $\beta \in \{0.0001, 0.001, 0.01, 0.1\}$ on UCLA and evaluating the effect of $\sigma$ on ADNI and REST-meta-MDD (Figure~\ref{fig:heatmap}).

\subsubsection{Joint sensitivity of $\beta_2$ and $\beta_3$ across different $\beta_1$ settings}

Figure~\ref{fig:heatmap} summarizes the joint sensitivity of $\beta_2$ and $\beta_3$ across different values
$\beta_1 \in \{0.0001, 0.001, 0.01, 0.1\}$ on UCLA. The best performance consistently occurs in the upper-right region with larger $\beta_2$ and $\beta_3$, provided that $\beta_1$ is moderately set. In particular, $\beta_1{=}0.01$ yields the optimal operating point at $\beta_2{=}0.1$ and $\beta_3{=}0.1$. When $\beta_1$ is too small, insufficient compression of the second-order view leaves noise and redundancy largely unregulated; when $\beta_1$ is too large, excessive regularization removes discriminative pairwise information, reducing accuracy. Overall, the heatmaps indicate a stable optimum characterized by moderate regularization for the pairwise view and strong regularization for higher-order views.

\subsubsection{Individual hyperparameter sensitivity}

Varying one coefficient at a time while fixing the other two at their optimal values shows that accuracy is maximized at $\beta_1{=}0.01$ and $\beta_2{=}\beta_3{=}0.1$, and degrades when any coefficient decreases by one or more orders of magnitude. The second-order view exhibits a moderate-regularization preference, where too little compression retains noisy correlations and too much compression discards useful connectivity patterns. In contrast, the higher-order views are more sensitive and benefit from stronger entropy regularization, because the third- and fourth-order $\mathcal{O}$-information tensors encode substantially higher-dimensional interaction spaces than pairwise graphs and are therefore more prone to overfitting.

\begin{table}[h]
\centering
\caption{Kernel bandwidth sensitivity analysis on ADNI and REST-meta-MDD datasets using 2-view configuration with randomized approximation.}
\label{tab:sigma_sensitivity}
\begin{tabular}{lcc}
\hline
\textbf{$\sigma$} & \textbf{ADNI} & \textbf{REST-meta-MDD} \\
\hline
0.1 & 70.44 $\pm$ 5.87 & 69.87 $\pm$ 3.62 \\
1 & 71.66 $\pm$ 5.21 & 71.31 $\pm$ 3.14 \\
5 & 72.94 $\pm$ 4.52 & 71.97 $\pm$ 2.83 \\
10 & 70.53 $\pm$ 5.79 & 71.53 $\pm$ 3.27 \\
Median & 73.16 $\pm$ 4.38 & 72.41 $\pm$ 2.71 \\
\hline
\end{tabular}
\end{table}

\subsubsection{Kernel bandwidth sensitivity}

We examine the kernel bandwidth $\sigma$ in the matrix-based R\'enyi entropy estimator by comparing fixed choices $\sigma \in \{0.1, 1, 5, 10\}$ with a data-driven median heuristic that sets $\sigma$ to the median pairwise Euclidean distance within each mini-batch. Table~\ref{tab:sigma_sensitivity} reports ADNI and REST-meta-MDD accuracy using the 2-view (pairwise + third-order) configuration with randomized $\mathcal{O}$-information computation. The median heuristic performs best on both datasets (73.16\% on ADNI and 72.41\% on REST-meta-MDD), outperforming all fixed bandwidths, while $\sigma{=}5$ is the strongest fixed setting. This suggests that adapting the kernel scale during training improves robustness by avoiding over-smoothing at large $\sigma$ and noise sensitivity at small $\sigma$, and provides a parameter-free alternative that generalizes across datasets and training dynamics.

\subsubsection{Summary of hyperparameter analysis}

Overall, the sensitivity analysis supports a hierarchical compression strategy with $\beta_1{=}0.01$ and $\beta_2{=}\beta_3{=}0.1$, corresponding to moderate regularization for the pairwise view and stronger regularization for higher-order views. Performance improves as $\beta_2$ and $\beta_3$ increase, consistent with the higher capacity and overfitting risk of third- and fourth-order tensors, while $\beta_1$ exhibits an intermediate optimum because overly weak regularization retains pairwise noise and overly strong regularization over-compresses discriminative connectivity. Despite these trends, accuracy is stable within a practical range, varying by less than 1 percentage point when $\beta_2,\beta_3 \in [0.01, 0.1]$ with $\beta_1{=}0.01$, indicating robustness to moderate hyperparameter misspecification and straightforward tuning via cross-validation. For the kernel bandwidth, the median heuristic consistently outperforms fixed $\sigma$, suggesting that adaptive kernel scaling further improves generalization.

\section{Discussion and Conclusion}

\subsection{Comparison with Previous MvHo-IB and Computational Efficiency}

Table~\ref{tab:efficiency} compares the performance and scalability of different entropy estimation strategies and model configurations. The original MvHo-IB framework employed an exhaustive computation of all third-order $\mathcal{O}$-information terms, which achieved excellent accuracy on smaller cohorts but became computationally prohibitive on larger datasets like REST and ABIDE.

\begin{table}[h]
\centering
\caption{Comparison of entropy estimation strategies and model configurations across four datasets. Accuracy (\%) is reported. ``Previous'' refers to the original MvHo-IB implementation; ``Gauss'' and ``Random'' denote the Gaussian analytical and randomized approximation-based estimators in MvHo-IB++, respectively.}
\label{tab:efficiency}
\resizebox{0.48\textwidth}{!}{%
\begin{tabular}{lcccc}
\hline
\textbf{Method} & \textbf{UCLA} & \textbf{ADNI} & \textbf{REST} & \textbf{ABIDE} \\
\hline
2-view (Previous) & 83.12 $\pm$ 5.74 & 73.23 $\pm$ 4.37 & -- & -- \\
2-view (Gauss) & 82.27 $\pm$ 5.86 & 72.54 $\pm$ 4.58 & 71.61 $\pm$ 2.73 & 72.17 $\pm$ 2.84 \\
2-view (Random) & 82.47 $\pm$ 5.71 & 72.96 $\pm$ 4.42 & 71.97 $\pm$ 2.68 & 71.93 $\pm$ 2.91 \\
3-view (Gauss) 4D & 82.94 $\pm$ 5.52 & 73.17 $\pm$ 4.31 & 72.06 $\pm$ 2.61 & 72.41 $\pm$ 2.76 \\
3-view (Random) 4D & 83.36 $\pm$ 5.63 & 73.41 $\pm$ 4.26 & 72.53 $\pm$ 2.54 & 72.61 $\pm$ 2.68 \\
\hline
\end{tabular}%
}
\end{table}

\subsubsection{Computational scalability and approximation accuracy}

The \textbf{2-view (Previous)} configuration corresponds to the original MvHo-IB framework, which models pairwise and third-order interactions via direct entropy computation without acceleration. It reaches 83.12\% on UCLA and 73.23\% on ADNI but is infeasible on the larger REST (1,604 subjects) and ABIDE (1,099 subjects) cohorts, as reflected by missing entries, due to the cubic cost of third-order $\mathcal{O}$-information computation and the memory required to store full covariance matrices.

Our acceleration strategies remove this bottleneck. \textit{2-view (Gauss)} uses Gaussian analytical approximations, achieving 82.27\% on UCLA and 72.54\% on ADNI with small drops of 0.85 and 0.69 points relative to \textit{2-view (Previous)}, while scaling to REST (71.61\%) and ABIDE (72.17\%). \textit{2-view (Random)} applies a randomized approximation for matrix-based R\'enyi entropy and achieves 82.47\% on UCLA, 72.96\% on ADNI, 71.97\% on REST, and 71.93\% on ABIDE, consistently exceeding \textit{2-view (Gauss)} on UCLA and ADNI and matching or improving it on the larger cohorts. Overall, both approximations introduce minimal error while enabling large-scale multi-site analysis.

\subsubsection{Wall-clock computation time}

To quantify the computational efficiency gains in absolute terms, we measured the wall-clock time required to compute all third-order and fourth-order $\mathcal{O}$-information tensors on the UCLA dataset (105 ICA networks) using a single NVIDIA A100 GPU. Table~\ref{tab:computational_time} reports the results.

\begin{table}[h]
\centering
\caption{Computational time comparison for $\mathcal{O}$-information computation on UCLA dataset (105 ICA networks, single A100 GPU).}
\label{tab:computational_time}
\begin{tabular}{lccc}
\hline
\textbf{Method} & \textbf{Time (min)} & \textbf{Speedup} & \textbf{Relative Cost} \\
\hline
Without & 5790.84 & $1.0\times$ & 100\% \\
Gaussian & 53.26 & $108.8\times$ & 0.92\% \\
Randomized & 170.31 & $34.0\times$ & 2.9\% \\
\hline
\end{tabular}
\end{table}

Without approximation, exhaustive computation requires 5790.84 minutes, rendering it impractical for iterative model development and large-scale studies. The Gaussian analytical approximation reduces computation time to 53.26 minutes, achieving a 108.8-fold speedup while requiring only 0.92\% of the original computational cost. The randomized approximation completes in 170.31 minutes with a 34.0-fold speedup and 2.9\% relative cost. Both strategies substantially exceed the 30-fold acceleration threshold, validating the computational efficiency claims in the abstract and enabling practical deployment of exhaustive higher-order interaction modeling on large cohorts.

\subsubsection{Benefits of fourth-order interactions}

Extending the framework to fourth-order interactions further improves performance. The \textit{3-view (Gauss) 4D} configuration adds a fourth-order $\mathcal{O}$-information tensor computed via Gaussian approximation, achieving 82.94\% on UCLA, 73.17\% on ADNI, 72.06\% on REST, and 72.41\% on ABIDE. Compared to 2-view (Gauss), the fourth-order view contributes gains of 0.67, 0.63, 0.45, and 0.24 percentage points, respectively. The \textit{3-view (Random) 4D} variant, which employs randomized approximation for all entropy computations, attains 83.36\% on UCLA, 73.41\% on ADNI, 72.53\% on REST, and 72.61\% on ABIDE, matching or exceeding the original MvHo-IB on UCLA and ADNI while extending coverage to the previously intractable REST and ABIDE datasets. The consistent superiority of the randomized approximation-based approach (0.42 points over Gauss on UCLA, 0.24 points on ADNI, 0.47 points on REST, 0.20 points on ABIDE) suggests that the randomized approximation provides more accurate entropy estimates than the Gaussian approximation, particularly when integrated across multiple views.

\subsubsection{Summary and implications}

The comparison reveals three key insights. First, the acceleration strategies, Gaussian analytical approximation and randomized approximation for matrix-based Rényi's entropy, enable scalable multi-order $\mathcal{O}$-information computation on large cohorts with minimal performance degradation (less than 1\% loss compared to the exhaustive original method). Second, the randomized approximation consistently outperforms Gaussian approximation, likely due to its nonparametric nature and robustness to deviations from Gaussianity in real fMRI data. Third, the addition of fourth-order interactions provides consistent, albeit modest, improvements across all datasets (0.24,0.67 points), confirming that higher-order brain network dependencies carry complementary diagnostic signal. Together, these findings establish \textsc{MvHo-IB++} as a computationally efficient and statistically principled framework for capturing multi-order brain interactions at scale.

\subsection{Interpretability: Discriminative Higher-Order Patterns via Grad-CAM}

\begin{figure}[!htbp]
\centering
\includegraphics[width=0.45\textwidth]{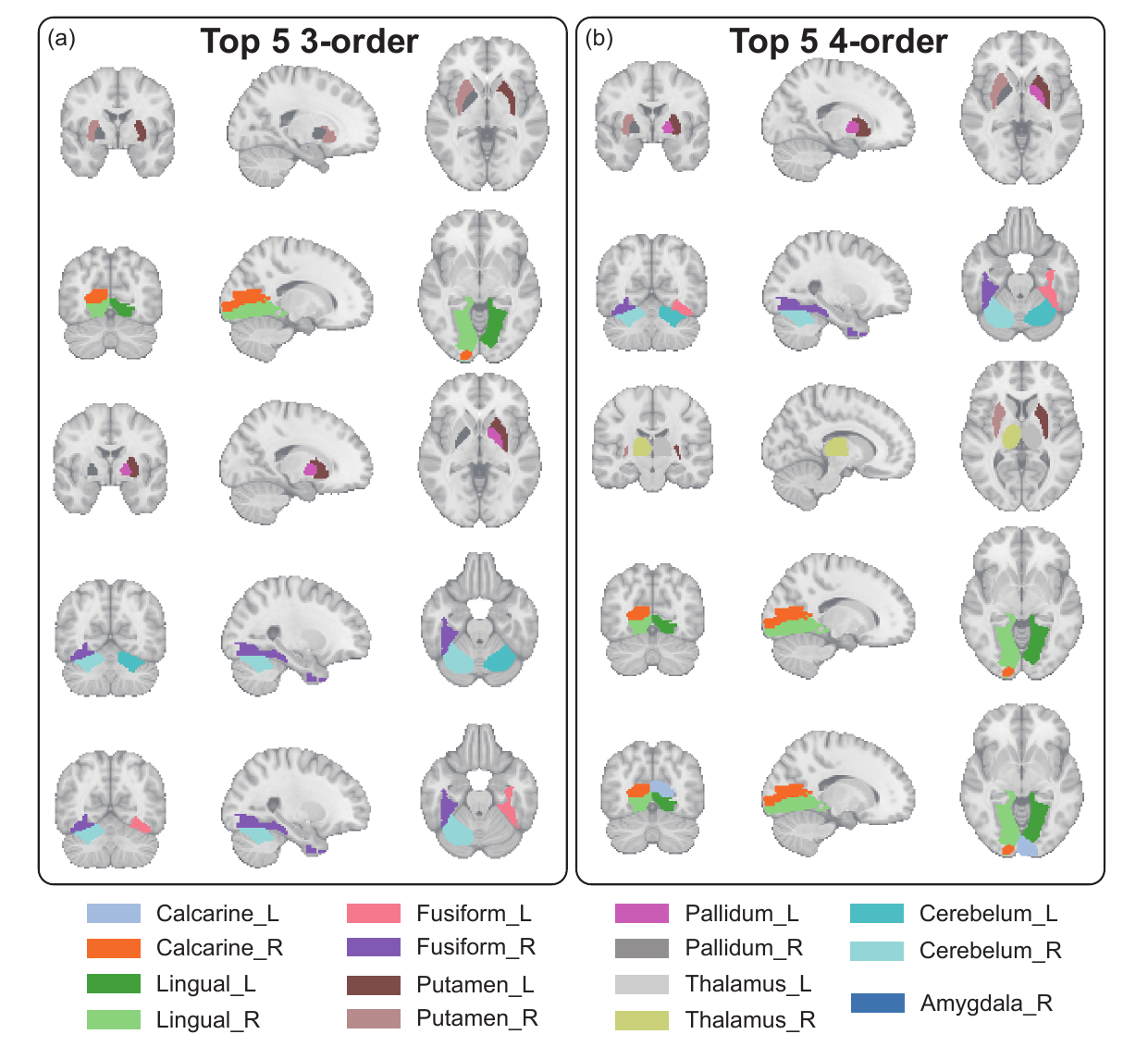}
\caption{Discriminative higher-order interaction patterns in major depressive disorder. (a) Top-5 third-order interactions. (b) Top-5 fourth-order interactions.}
\label{fig:mdd}
\end{figure}

To identify which higher-order interactions contribute most to diagnostic classification, we apply Gradient-weighted Class Activation Mapping (Grad-CAM)~\cite{selvaraju2017grad} to the learned third- and fourth-order $\mathcal{O}$-information tensors. Grad-CAM computes the gradient of the class prediction with respect to each tensor element, highlighting brain region/network combinations whose interaction patterns are most discriminative. We identify the top-5 triplets and quadruplets with the highest absolute gradient magnitudes across the UCLA, REST-meta-MDD, and ADNI datasets, then compare their mean $\mathcal{O}$-information values between patient and healthy control groups. Table~\ref{tab:mdd} and Figure~\ref{fig:mdd} present the results for major depressive disorder as a representative example; detailed results for all three disorders are provided in the Supplementary Material.

\begin{table}[h]
\centering
\caption{Top-5 discriminative third-order and fourth-order interactions in REST-meta-MDD dataset identified by Grad-CAM. AAL region indices are shown.}
\label{tab:mdd}
\resizebox{0.5\textwidth}{!}{%
\begin{tabular}{lccc}
\hline
\textbf{Triplet/Quadruplet} & \textbf{MDD (Mean $\pm$ SD)} & \textbf{HC (Mean $\pm$ SD)} & \textbf{$\Delta$ (MDD - HC)} \\
\hline
73, 74, 76 & $0.211 \pm 0.161$ & $0.149 \pm 0.121$ & +0.063 \\
44, 47, 48 & $0.360 \pm 0.228$ & $0.304 \pm 0.229$ & +0.056 \\
73, 75, 76 & $0.183 \pm 0.150$ & $0.128 \pm 0.110$ & +0.055 \\
56, 99, 100 & $0.117 \pm 0.127$ & $0.066 \pm 0.104$ & +0.051 \\
55, 56, 100 & $0.074 \pm 0.111$ & $0.032 \pm 0.083$ & +0.042 \\
\hline
\hline
73, 74, 75, 76 & $0.268 \pm 0.179$ & $0.187 \pm 0.142$ & +0.081 \\
55, 56, 99, 100 & $0.143 \pm 0.141$ & $0.079 \pm 0.108$ & +0.064 \\
73, 74, 77, 78 & $0.232 \pm 0.173$ & $0.161 \pm 0.129$ & +0.071 \\
44, 47, 48, 42 & $0.408 \pm 0.238$ & $0.344 \pm 0.231$ & +0.064 \\
43, 44, 47, 48 & $0.397 \pm 0.233$ & $0.334 \pm 0.228$ & +0.063 \\
\hline
\end{tabular}%
}
\end{table}

Across all three disorders, Grad-CAM analysis reveals disorder-specific higher-order connectivity signatures. In schizophrenia, the most discriminative interactions involve cerebellar--occipitotemporal and thalamic circuits, with both increased and decreased $\mathcal{O}$-information indicating imbalanced rather than uniformly disrupted connectivity. In major depressive disorder (Table~\ref{tab:mdd}, Figure~\ref{fig:mdd}), the top patterns implicate striato-pallido-thalamic, visual-limbic, and fusiform-cerebellar circuits, all showing hyperconnectivity. In mild cognitive impairment, frontal executive, right-lateralized visual, and cerebellar cognitive regions exhibit elevated higher-order interactions, consistent with compensatory network rigidification in prodromal Alzheimer's disease. See the Supplementary Material for detailed neuroanatomical interpretation.

Notably, many discriminative fourth-order quadruplets directly extend their corresponding third-order triplets across all three disorders, indicating that the identified regions participate in coherent multi-way functional dependencies rather than spurious combinations. This hierarchical consistency across interaction orders confirms that the third-order $\mathcal{O}$-information captures local synergistic triads, while the fourth-order $\mathcal{O}$-information extends these to broader, more distributed interaction motifs. The convergent yet disorder-specific neuroanatomical findings validate the complementary nature of the multi-view framework.

In this work, we introduced MvHo-IB++, a scalable multi-view information bottleneck framework for psychiatric diagnosis that systematically integrates pairwise, triadic, and tetradic brain interactions. By combining $\mathcal{O}$-information--based higher-order modeling with two independent acceleration strategies, we made exhaustive third- and fourth-order interaction enumeration computationally feasible on large multisite cohorts. Across extensive experiments on four benchmark datasets (UCLA, ADNI, REST-meta-MDD, and ABIDE), MvHo-IB++ consistently achieves state-of-the-art performance, outperforming graph neural networks, information bottleneck variants, and representative hypergraph-based approaches. Our results suggest that systematic enumeration of higher-order interactions with redundancy compression yields strong and stable diagnostic performance while avoiding heuristic hyperedge construction and pruning strategies that may introduce selection bias or miss informative interaction patterns.

Beyond quantitative gains, MvHo-IB++ reveals discriminative higher-order connectivity signatures that are not explicitly accessible to conventional hypergraph formulations. By explicitly characterizing redundancy- and synergy-dominated dependencies, the proposed framework identifies disorder specific multi-region biomarkers that emerge only at triadic and tetradic levels. Extensive ablation studies further confirm that each component contributes synergistically and that performance remains stable across a broad range of regularization coefficients and kernel parameters, demonstrating robustness to hyperparameter specification. More broadly, the proposed framework offers a general route for integrating explicit higher-order interaction measures with multi-view representation learning in structured biomedical signals.



\bibliographystyle{elsarticle-num-names} 
\bibliography{ref}
\newpage
\appendix
\section*{A. Matrix-Based R\'enyi's $\alpha$-Order Entropy Functional for the IB Terms}

\subsection*{A.1 Objective Context}
In the multi-view IB objective, the compression terms are
\begin{equation}
\beta_1 H(Z_1)+\beta_2 H(Z_2)+\beta_3 H(Z_3),
\end{equation}
where $Z_1,Z_2,Z_3$ are latent representations from the three encoders.
This section details how each $H(Z_v)$ is estimated with a matrix-based R\'enyi entropy functional on mini-batches.

\subsection*{A.2 Mini-Batch Kernel Matrix Construction}
For each view-specific latent representation $Z_1$, $Z_2$, or $Z_3$, let a mini-batch contain
$N_b$ latent vectors $\{z_1,\ldots,z_{N_b}\}$, where $N_b \le N$ and $N$ is the number of subjects in the dataset.
Define the Gaussian kernel
\begin{equation}
\kappa(z_i,z_j)=\exp\!\left(-\frac{\lVert z_i-z_j\rVert_2^2}{2\sigma^2}\right),
\quad i,j=1,\ldots,N_b.
\end{equation}
Construct the Gram matrix $\mathbf{K}\in\mathbb{R}^{N_b\times N_b}$ with
$\mathbf{K}_{ij}=\kappa(z_i,z_j)$, then normalize by trace:
\begin{equation}
\mathbf{G}=\frac{\mathbf{K}}{\mathrm{tr}(\mathbf{K})}.
\end{equation}
Thus $\mathbf{G}$ is positive semidefinite and satisfies $\mathrm{tr}(\mathbf{G})=1$.

\subsection*{A.3 Matrix-Based R\'enyi Entropy Estimator}
For order $\alpha>0$, $\alpha\neq 1$, define
\begin{equation}
\widehat{H}_{\alpha}(Z)
=\frac{1}{1-\alpha}\log\!\left(\mathrm{tr}\!\left(\mathbf{G}^{\alpha}\right)\right).
\end{equation}
In practice, integer $\alpha$ is commonly used.
This avoids explicit density estimation and can be directly optimized by backpropagation through the kernel matrix.

\subsection*{A.4 Three-View Entropy Terms in Training}
For each training mini-batch, compute:
\begin{align}
\widehat{H}_{\alpha}(Z_1)&=\frac{1}{1-\alpha}\log\!\left(\mathrm{tr}\!\left(\mathbf{G}_1^{\alpha}\right)\right),\\
\widehat{H}_{\alpha}(Z_2)&=\frac{1}{1-\alpha}\log\!\left(\mathrm{tr}\!\left(\mathbf{G}_2^{\alpha}\right)\right),\\
\widehat{H}_{\alpha}(Z_3)&=\frac{1}{1-\alpha}\log\!\left(\mathrm{tr}\!\left(\mathbf{G}_3^{\alpha}\right)\right).
\end{align}
The implemented loss term is
\begin{equation}
\mathcal{L}_{\text{ent}}
=\beta_1 \widehat{H}_{\alpha}(Z_1)
+\beta_2 \widehat{H}_{\alpha}(Z_2)
+\beta_3 \widehat{H}_{\alpha}(Z_3).
\end{equation}

\subsection*{A.5 Algorithmic Summary}
\begin{algorithm}[h]
\caption{Mini-Batch Entropy Estimation for $H(Z_1),H(Z_2),H(Z_3)$}
\begin{algorithmic}[1]
\REQUIRE Mini-batch latent sets $\{z_i\}_{i=1}^{N_b}$ from each of $Z_1,Z_2,Z_3$; $\alpha$; $\sigma$; $\beta_1,\beta_2,\beta_3$
\FOR{$v=1$ to $3$}
\STATE Build $\mathbf{K}_v$ using $\kappa(z_i,z_j)=\exp(-\|z_i-z_j\|_2^2/(2\sigma^2))$
\STATE Normalize $\mathbf{G}_v \leftarrow \mathbf{K}_v/\mathrm{tr}(\mathbf{K}_v)$
\STATE Compute $\widehat{H}_{\alpha}(Z_v)=\frac{1}{1-\alpha}\log(\mathrm{tr}(\mathbf{G}_v^\alpha))$
\ENDFOR
\STATE Return $\mathcal{L}_{\text{ent}}=\beta_1\widehat{H}_{\alpha}(Z_1)+\beta_2\widehat{H}_{\alpha}(Z_2)+\beta_3\widehat{H}_{\alpha}(Z_3)$
\end{algorithmic}
\end{algorithm}

\section*{B. Detailed Derivations and Complexity Analysis}

\subsection*{B.1 Gaussian Analytical Approximation}
When the joint distribution of the involved variables is well-approximated as multivariate Gaussian, differential entropy is available in closed form via the covariance matrix $\Sigma \in \mathbb{R}^{K \times K}$~\cite{gaussOinfo2310}:
\begin{equation}
H(X_1,\ldots,X_K)
= \frac{1}{2}\log\!\big((2\pi e)^K \det(\Sigma)\big).
\end{equation}

The total correlation (TC) is first defined as:
\begin{equation}
\mathrm{TC}(X_1,\ldots,X_K)=\sum_{i=1}^{K}H(X_i)-H(X_1,\ldots,X_K).
\end{equation}
Under Gaussianity, this reduces to:
\begin{equation}
\mathrm{TC}(X_1,\ldots,X_K)=\frac{1}{2}\log\!\left(\frac{\det\!\big(\mathrm{diag}(\Sigma)\big)}{\det(\Sigma)}\right).
\end{equation}

The dual total correlation (DTC) is first defined as:
\begin{equation}
\mathrm{DTC}(X_1,\ldots,X_K)=H(X_1,\ldots,X_K)-\sum_{i=1}^{K}H(X_i\mid X_{[K]\backslash i}),
\end{equation}
where $X_{[K]\backslash i}$ denotes all variables except $X_i$.
Under the Gaussian assumption,
\begin{equation}
H(X_i \mid X_{[K]\backslash i})
= \frac{1}{2}\log\!\big((2\pi e)\,\sigma^2_{i\mid [K]\backslash i}\big),
\end{equation}
and equivalently:
\begin{equation}
\mathrm{DTC}(X_1,\ldots,X_K)=\sum_{i=1}^{K}H(X_{[K]\backslash i})-(K-1)H(X_1,\ldots,X_K).
\end{equation}
Under Gaussianity, each term $H(X_{[K]\backslash i})$ is computed from the corresponding $(K{-}1)\times(K{-}1)$ principal minor of $\Sigma$. The $\mathcal{O}$-information is then obtained as:
\begin{equation}
\mathcal{O}(X_1,\ldots,X_K)=\mathrm{TC}(X_1,\ldots,X_K)-\mathrm{DTC}(X_1,\ldots,X_K).
\end{equation}
This closed-form computation requires only covariance determinants and principal minors (computed from the upper triangle), enabling efficient $\mathcal{O}$-information estimation for $K=3$ or $K=4$ without density estimation. Given $T$ time points and $C$ brain regions/networks, the $K \times K$ covariance matrix $\Sigma \in \mathbb{R}^{K \times K}$ is computed in $\mathcal{O}(T \cdot K^2)$ for each $K$-tuple, and the determinant computation requires $\mathcal{O}(K^3)$. For all ${C \choose K}$ combinations, the total complexity is $\mathcal{O}({C \choose K} \cdot (T \cdot K^2 + K^3))$, which simplifies to $\mathcal{O}(C^K \cdot T)$ for small $K$.

\subsection*{B.2 Randomized Approximation for Matrix-Based R\'enyi's Entropy}
For non-Gaussian settings, we estimate entropies with the matrix-based R\'enyi's $\alpha$-order functional. Given $T$ time-point samples $\{\mathbf{x}_1,\ldots,\mathbf{x}_T\}$ and the normalized Gram matrix $\mathbf{G}\in\mathbb{R}^{T\times T}$ built with a Gaussian kernel $\kappa(\mathbf{x}_i,\mathbf{x}_j)=\exp(-\|\mathbf{x}_i-\mathbf{x}_j\|^2/(2\sigma^2))$, the entropy is
\begin{equation}
H_{\alpha}(\mathbf{X})=\frac{1}{1-\alpha}\log\!\big(\mathrm{tr}(\mathbf{G}^{\alpha})\big).
\end{equation}

We avoid explicit eigendecomposition by randomized trace estimation: draw $s$ i.i.d. standard Gaussian probe vectors $\{\mathbf{g}_i\}_{i=1}^s$ and compute
\begin{equation}
\widetilde{H}_{\alpha}(\mathbf{X})=\frac{1}{1-\alpha}\log\!\left(\frac{1}{s}\sum_{i=1}^{s}\mathbf{g}_i^{\top}\mathbf{G}^{\alpha}\mathbf{g}_i\right).
\end{equation}
The complete randomized approximation procedure is summarized in Algorithm~\ref{alg:renyi_approx}~\cite{random}.

\begin{algorithm}[h]
\caption{Integer order matrix-based R\'enyi's entropy estimation}
\label{alg:renyi_approx}
\begin{algorithmic}[1]
\REQUIRE Kernel matrix $\mathbf{G} \in \mathbb{R}^{T \times T}$, number of random vectors $s$, integer order $\alpha \geq 2$
\ENSURE Approximation to $H_{\alpha}(\mathbf{G})$
\STATE Generate $s$ independent random standard Gaussian vectors $\mathbf{g}_i$, $i = 1, \ldots, s$
\STATE \textbf{return} $\widetilde{H}_{\alpha}(\mathbf{G}) = \frac{1}{1-\alpha} \log\!\left( \frac{1}{s}\sum_{i=1}^{s} \mathbf{g}_i^{\top} \mathbf{G}^{\alpha} \mathbf{g}_i \right)$
\end{algorithmic}
\end{algorithm}
Powers of $\mathbf{G}$ act through repeated matrix--vector products, enabling a complexity on the order of $\mathcal{O}(\alpha\,s\,T^2)$ and matrix storage restricted to the upper triangle. Using this randomized entropy estimator, we compute TC and DTC as:
\begin{align}
\mathrm{TC}_{\alpha}(X_1,\ldots,X_K)&=\sum_{i=1}^{K}\widetilde{H}_{\alpha}(X_i)-\widetilde{H}_{\alpha}(X_1,\ldots,X_K),\\
\mathrm{DTC}_{\alpha}(X_1,\ldots,X_K)&=\sum_{i=1}^{K}\widetilde{H}_{\alpha}(X_{[K]\backslash i})-(K-1)\widetilde{H}_{\alpha}(X_1,\ldots,X_K),
\end{align}
where $\widetilde{H}_{\alpha}(X_i)$ is estimated from the Gram matrix of the $i$-th variable, $\widetilde{H}_{\alpha}(X_1,\ldots,X_K)$ from the joint Gram matrix constructed via kernel product, and $\widetilde{H}_{\alpha}(X_{[K]\backslash i})$ from the joint Gram matrix excluding the $i$-th variable. The $\mathcal{O}$-information is then obtained as
\begin{equation}
\mathcal{O}_{\alpha}=\mathrm{TC}_{\alpha}-\mathrm{DTC}_{\alpha},
\end{equation}
providing scalable estimates of third- and fourth-order interactions without distributional assumptions. The Gram matrix construction requires $\mathcal{O}(T^2)$ for $T$ time points, and each randomized trace estimation costs $\mathcal{O}(\alpha \cdot s \cdot T^2)$ where $s$ is the number of probe vectors. Computing $\mathcal{O}$-information for one $K$-tuple requires $(2K+1)$ entropy evaluations. For all ${C \choose K}$ combinations, the total complexity is $\mathcal{O}({C \choose K} \cdot K \cdot \alpha \cdot s \cdot T^2)$, which simplifies to $\mathcal{O}(C^K \cdot T^2)$ for fixed $\alpha$ and $s$.

\section*{C. Detailed Neuroanatomical Interpretation}

To complement the summary in the main text, this section provides the full Grad-CAM--based neuroanatomical interpretation of discriminative higher-order interaction patterns for each disorder.

\subsection*{C.1 Schizophrenia}

Table~\ref{tab:sz} presents the top-5 discriminative third-order and fourth-order interactions in the UCLA schizophrenia dataset.

\begin{table}[h]
\centering
\caption{Top-5 discriminative third-order and fourth-order interactions in UCLA schizophrenia dataset identified by Grad-CAM. ICA105 component indices are shown.}
\label{tab:sz}
\resizebox{0.5\textwidth}{!}{%
\begin{tabular}{lccc}
\hline
\textbf{Triplet/Quadruplet} & \textbf{SZ (Mean $\pm$ SD)} & \textbf{HC (Mean $\pm$ SD)} & \textbf{$\Delta$ (SZ - HC)} \\
\hline
6, 14, 16 & $0.368 \pm 0.248$ & $0.237 \pm 0.227$ & +0.131 \\
2, 68, 71 & $0.175 \pm 0.263$ & $0.066 \pm 0.171$ & +0.109 \\
41, 42, 44 & $0.367 \pm 0.236$ & $0.264 \pm 0.183$ & +0.103 \\
1, 2, 71 & $0.195 \pm 0.207$ & $0.098 \pm 0.150$ & +0.097 \\
83, 84, 96 & $0.220 \pm 0.200$ & $0.316 \pm 0.220$ & -0.096 \\
\hline
\hline
4, 6, 14, 16 & $0.412 \pm 0.263$ & $0.251 \pm 0.228$ & +0.161 \\
2, 68, 71, 72 & $0.218 \pm 0.274$ & $0.084 \pm 0.179$ & +0.134 \\
41, 42, 43, 44 & $0.403 \pm 0.247$ & $0.276 \pm 0.192$ & +0.127 \\
1, 2, 71, 72 & $0.247 \pm 0.223$ & $0.113 \pm 0.158$ & +0.134 \\
83, 84, 95, 96 & $0.208 \pm 0.211$ & $0.342 \pm 0.229$ & -0.134 \\
\hline
\end{tabular}%
}
\end{table}

In schizophrenia, \textsc{MvHo-IB++} identifies three distinct pathological systems (Table~\ref{tab:sz}).

The first system involves the triplet (6, 14, 16), connecting the networks cerebellar (ICA6), occipitotemporal (ICA14, ICA16). This connection shows the strongest abnormality, with elevated $\mathcal{O}$-information in SZ ($0.368 \pm 0.248$) compared to HC ($0.237 \pm 0.227$, $\Delta = +0.131$). Adding the network cerebellar (ICA4) to form the quadruplet (4, 6, 14, 16) reveals an even greater dysfunction ($\Delta = +0.161$).

The second system includes the triplet (41, 42, 44), spanning the networks extended thalamic (ICA41, ICA42 and ICA44). SZ shows higher interactions ($0.367 \pm 0.236$) than HC ($0.264 \pm 0.183$, $\Delta = +0.103$), which further increase in the quadruplet (41, 42, 43, 44; $\Delta = +0.127$). This connection reflects corollary discharge dysfunction.

In contrast, the third system, comprising the triplet (83, 84, 96) and quadruplet (83, 84, 95, 96), links the networks temporoparietal (ICA83, ICA84), and default mode (ICA96). Here, $\mathcal{O}$-information is reduced in SZ (triplet: $\Delta = -0.096$; quadruplet: $\Delta = -0.134$), the only negative-valued pattern among the top discriminators.

Together, these findings support the dysconnection hypothesis~\cite{jiang2024neuroimaging}, indicating that schizophrenia stems not from uniform network dysfunction but from imbalanced connectivity across brain networks.

\subsection*{C.2 Major Depressive Disorder}

In major depressive disorder, \textsc{MvHo-IB++} identifies three distinct pathological systems.

The first system involves the triplet (73, 74, 76), connecting bilateral putamen (AAL73, AAL74) and right pallidum (AAL76). This connection shows the strongest abnormality, with elevated $\mathcal{O}$-information in MDD ($0.211 \pm 0.161$) compared to HC ($0.149 \pm 0.121$, $\Delta = +0.063$). Adding the left pallidum (AAL75) to form the quadruplet (73, 74, 75, 76) reveals the greatest dysfunction ($\Delta = +0.081$). This basal ganglia circuit also extends to the thalamus in the quadruplet (73, 74, 77, 78; $\Delta = +0.071$), reflecting hyperconnectivity within the striato-pallido-thalamic pathway.

The second system includes the triplet (44, 47, 48), spanning right calcarine cortex (AAL44) and bilateral lingual gyrus (AAL47, AAL48). MDD shows higher interactions ($0.360 \pm 0.228$) than HC ($0.304 \pm 0.229$, $\Delta = +0.056$), which further increase in the quadruplets extending bilaterally (43, 44, 47, 48; $\Delta = +0.063$) and incorporating the amygdala (44, 47, 48, 42; $\Delta = +0.064$). This visual-limbic circuit reflects aberrant connectivity between early visual processing regions and emotional centers.

The third system, comprising the triplets (56, 99, 100) and (55, 56, 100), links bilateral fusiform gyrus (AAL55, AAL56) with bilateral cerebellum lobule VI (AAL99, AAL100). The combined quadruplet (55, 56, 99, 100) shows elevated $\mathcal{O}$-information ($\Delta = +0.064$), indicating dysconnection between ventral visual stream and the cerebellar cognitive control regions.

Together, these findings reveal that major depressive disorder involves hyperconnectivity across multiple systems: striato-pallidal motivational circuits, visual-limbic processing pathways, and fusiform-cerebellar integration networks, supporting the hypothesis that depression stems from imbalanced functional integration rather than localized deficits~\cite{sheline2009default, wise2017instability}.

\subsection*{C.3 Mild Cognitive Impairment}

\begin{figure}[!htbp]
\centering
\includegraphics[width=0.5\textwidth]{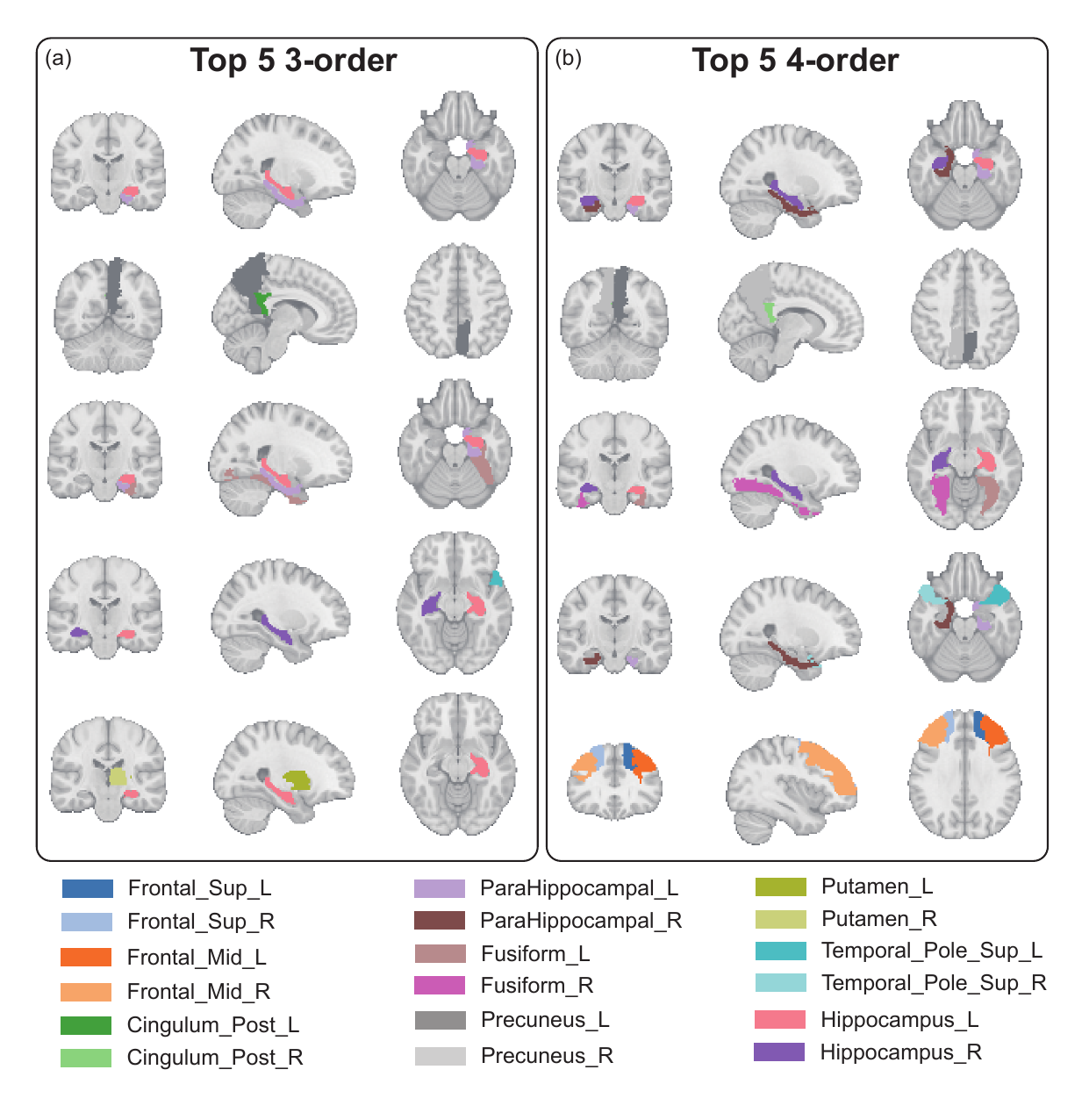}
\caption{Discriminative higher-order interaction patterns in mild cognitive impairment (ADNI dataset). (a) Top-5 third-order interactions. (b) Top-5 fourth-order interactions.}
\label{fig:adni}
\end{figure}

Table~\ref{tab:mci} and Figure~\ref{fig:adni} present the top-5 discriminative interactions and their spatial distribution in the ADNI dataset (MCI vs.\ NC).

\begin{table}[h]
\centering
\caption{Top-5 discriminative third-order and fourth-order interactions in ADNI dataset (MCI vs. NC) identified by Grad-CAM. AAL region indices are shown.}
\label{tab:mci}
\resizebox{0.5\textwidth}{!}{%
\begin{tabular}{lccc}
\hline
\textbf{Triplet/Quadruplet} & \textbf{MCI (Mean $\pm$ SD)} & \textbf{HC (Mean $\pm$ SD)} & \textbf{$\Delta$ (MCI - HC)} \\
\hline
3, 7, 22 & $0.362 \pm 0.091$ & $0.310 \pm 0.082$ & +0.052 \\
7, 22, 23 & $0.368 \pm 0.085$ & $0.321 \pm 0.076$ & +0.047 \\
3, 7, 23 & $0.429 \pm 0.103$ & $0.386 \pm 0.094$ & +0.043 \\
92, 100, 102 & $-0.002 \pm 0.089$ & $-0.040 \pm 0.085$ & +0.038 \\
48, 50, 52 & $0.319 \pm 0.088$ & $0.285 \pm 0.081$ & +0.034 \\
\hline
\hline
3, 7, 22, 23 & $0.467 \pm 0.112$ & $0.406 \pm 0.098$ & +0.061 \\
3, 4, 7, 8 & $0.403 \pm 0.096$ & $0.347 \pm 0.090$ & +0.056 \\
48, 50, 52, 54 & $0.356 \pm 0.094$ & $0.305 \pm 0.086$ & +0.051 \\
92, 100, 102, 104 & $0.010 \pm 0.098$ & $-0.035 \pm 0.091$ & +0.045 \\
7, 22, 23, 24 & $0.371 \pm 0.089$ & $0.332 \pm 0.083$ & +0.039 \\
\hline
\end{tabular}%
}
\end{table}

In mild cognitive impairment, \textsc{MvHo-IB++} identifies three distinct pathological systems (Figure~\ref{fig:adni}, Table~\ref{tab:mci}).

The first system involves the triplet (3, 7, 23), connecting left superior frontal gyrus (AAL3), left middle frontal gyrus (AAL7), and left medial superior frontal gyrus (AAL23). This connection shows elevated $\mathcal{O}$-information in MCI ($0.429 \pm 0.103$) compared to HC ($0.386 \pm 0.094$, $\Delta = +0.043$). Extending to the quadruplet (3, 7, 22, 23) by adding right olfactory cortex (AAL22) reveals the strongest dysfunction ($\Delta = +0.061$). This frontal-olfactory circuit further extends bilaterally in the quadruplet (3, 4, 7, 8; $\Delta = +0.056$), spanning bilateral superior and middle frontal gyri, reflecting compensatory hyperconnectivity within the prefrontal executive network.

The second system includes the triplet (48, 50, 52), spanning right lingual gyrus (AAL48), right superior occipital gyrus (AAL50), and right middle occipital gyrus (AAL52). MCI shows higher interactions ($0.319 \pm 0.088$) than HC ($0.285 \pm 0.081$, $\Delta = +0.034$), which further increase in the quadruplet (48, 50, 52, 54; $\Delta = +0.051$) by incorporating right inferior occipital gyrus (AAL54). This right-lateralized visual pathway dysfunction reflects impaired visuospatial processing characteristic of prodromal Alzheimer's disease.

The third system, comprising the triplet (92, 100, 102) and quadruplet (92, 100, 102, 104), links right cerebellar Crus I (AAL92), right cerebellar lobule VI (AAL100), right cerebellar lobule VIIb (AAL102), and right cerebellar lobule VIII (AAL104). MCI shows elevated $\mathcal{O}$-information (triplet: $\Delta = +0.038$; quadruplet: $\Delta = +0.045$), indicating dysconnection within the right cerebellar cognitive control regions that extend beyond motor coordination to support executive function and working memory.

Together, these findings reveal that mild cognitive impairment involves hyperconnectivity across frontal executive networks, visual processing pathways, and cerebellar cognitive regions, supporting the hypothesis that prodromal Alzheimer's disease reflects compensatory network rigidification rather than uniform disconnection.

\subsection*{C.4 Consistency and Complementarity of Multi-Order Patterns}

Many discriminative fourth-order quadruplets directly extend their corresponding third-order triplets. In schizophrenia, triplet (6, 14, 16) extends to quadruplet (4, 6, 14, 16), triplet (41, 42, 44) to quadruplet (41, 42, 43, 44), and triplet (83, 84, 96) to quadruplet (83, 84, 95, 96). Similarly, in MDD, triplet (73, 74, 76) extends to quadruplet (73, 74, 75, 76), and triplet (55, 56, 100) to quadruplet (55, 56, 99, 100). In MCI, triplet (3, 7, 23) extends to quadruplet (3, 7, 22, 23), triplet (48, 50, 52) to quadruplet (48, 50, 52, 54), and triplet (92, 100, 102) to quadruplet (92, 100, 102, 104). This hierarchical consistency across interaction orders indicates that the identified regions participate in coherent multi-way functional dependencies rather than spurious combinations. The third-order $\mathcal{O}$-information captures local synergistic triads, while the fourth-order $\mathcal{O}$-information extends these to broader, more distributed interaction motifs.

The convergent neuroanatomical findings across all three disorders, each showing disorder-specific patterns (cerebellar-visual-thalamic in SZ, striato-pallidal-visual-cerebellar in MDD, and frontal-occipital-cerebellar in MCI), validate the complementary nature of the multi-view framework and demonstrate that \textsc{MvHo-IB++} captures interpretable, clinically relevant higher-order connectivity signatures that align with established neuroscience knowledge of psychiatric and neurodegenerative pathophysiology.

\section*{D. Implementation Details of Compared Hypergraph Baselines}

To ensure fair comparison, we implemented the hypergraph construction procedures of HL~\cite{HL} and CcSi-MHAHGEL~\cite{CcSiMHAHGEL} following their original descriptions.

For HL, we constructed a shared cohort-level hypergraph using a $K$-nearest-neighbor (K-NN) scheme applied to the average functional connectivity matrix. Specifically, similarity between node features $x_i$ and $x_j$ was computed using a Gaussian kernel:
\begin{equation}
\mathrm{Sim}(v_i, v_j)
=
\exp\left(
-\frac{\|x_i - x_j\|_2^2}{\sigma^2}
\right).
\end{equation}
Each hyperedge consisted of a vertex and its $K-1$ nearest neighbors, forming a shared incidence matrix across subjects.

For CcSi-MHAHGEL, we constructed subject-specific hypergraphs via sparse linear regression. For each ROI $i$, its time series $x_i$ was modeled as a sparse combination of all other ROIs by solving:
\begin{equation}
\min_{\boldsymbol{\alpha}_i}
\frac{1}{2}\|x_i - A_i \boldsymbol{\alpha}_i\|_2^2
+
\lambda \|\boldsymbol{\alpha}_i\|_1,
\end{equation}
where $A_i$ contains the time series of all other ROIs. Hyperedges were formed by grouping ROI $i$ with regions corresponding to strictly positive regression coefficients.

For HyBRiD~\cite{Hybrid}, we adopted the learnable end-to-end hypergraph constructor as described in the original work, in which the hypergraph topology is parameterized and optimized via backpropagation.

All hypergraph-based models were implemented using a common HyperAtten backbone~\cite{gao2022hgnn+} to ensure architectural consistency across methods.

\end{document}